%% file: acl_latex.tex
\documentclass[11pt]{article}

\usepackage[final]{acl}

\usepackage{times}
\usepackage{latexsym}
\newcommand{\cmark}{\ding{51}}  % or \checkmark
\newcommand{\xmark}{\ding{55}}  % or \times
\usepackage{pifont}
\usepackage{colortbl}
\usepackage{amsmath}
\usepackage{xspace} 
\usepackage[utf8]{inputenc} % allow utf-8 input
\usepackage[T1]{fontenc}    % use 8-bit T1 fonts
\usepackage{hyperref}       % hyperlinks
\usepackage{url}            % simple URL typesetting
\usepackage{booktabs}       % professional-quality tables
\usepackage{amsfonts}       % blackboard math symbols
\usepackage{nicefrac}       % compact symbols for 1/2, etc.
\usepackage{microtype}      % microtypography
\usepackage{xcolor}         % colors
\usepackage{graphicx}
\usepackage{multirow}
\usepackage{moresize}
\usepackage{wrapfig}
\usepackage{tablefootnote}
\usepackage{float}
\usepackage{fontawesome5} % For the GitHub icon
\usepackage{academicons}   % For other academic/social icons (though you might need to add a custom one for Hugging Face)
\usepackage{hyperref}    % For creating clickable links
\usepackage{subcaption} % Add this in your preamble if not already present

\usepackage[T1]{fontenc}
\usepackage[utf8]{inputenc}

\usepackage{microtype}

\usepackage{inconsolata}
\usepackage{xspace}
\usepackage{graphicx}

\newcommand*{\AlgName}{\textsc{RUST-BENCH}\@\xspace}
\definecolor{ao(english)}{rgb}{0.0, 0.5, 0.0}

\title{\AlgName: Benchmarking LLM Reasoning on Unstructured Text within Structured Tables}

\author{Nikhil Abhyankar\textsuperscript{\tiny 1}, Purvi Chaurasia\textsuperscript{\tiny 2}, Sanchit Kabra\textsuperscript{\tiny 1}, Ananya Srivastava\textsuperscript{\tiny 2}, \\ \textbf{Vivek Gupta}\textsuperscript{\tiny 3}, \textbf{Chandan K. Reddy}\textsuperscript{\tiny 1} \\
        \textsuperscript{\tiny 1}Virginia Tech, \textsuperscript{\tiny 2}IGDTUW New Delhi, \textsuperscript{\tiny 3}Arizona State University \\
        }

\begin{document}
\maketitle
\begin{abstract}
Existing tabular reasoning benchmarks mostly test models on small, uniform tables, underrepresenting the complexity of real-world data and giving an incomplete view of Large Language Models' (LLMs) reasoning abilities. Real tables are long, heterogeneous, and domain-specific—mixing structured fields with free text and requiring multi-hop reasoning across thousands of tokens. To address this gap, we introduce \AlgName, a benchmark of 7,966 questions from 2,031 real-world tables spanning two domains: (i) RB-Science (NSF grant records) and (ii) RB-Sports (NBA statistics). Unlike prior work, \AlgName evaluates LLMs jointly across scale, heterogeneity, domain specificity, and reasoning complexity. Experiments with open-source and proprietary models show that LLMs struggle with heterogeneous schemas and complex multi-hop inference, revealing persistent weaknesses in current architectures and prompting strategies. \AlgName establishes a challenging new testbed for advancing tabular reasoning research.\footnote{ Correspondence: nikhilsa@vt.edu, vgupt140@asu.edu}
\\
{~\includegraphics[height=0.65em]{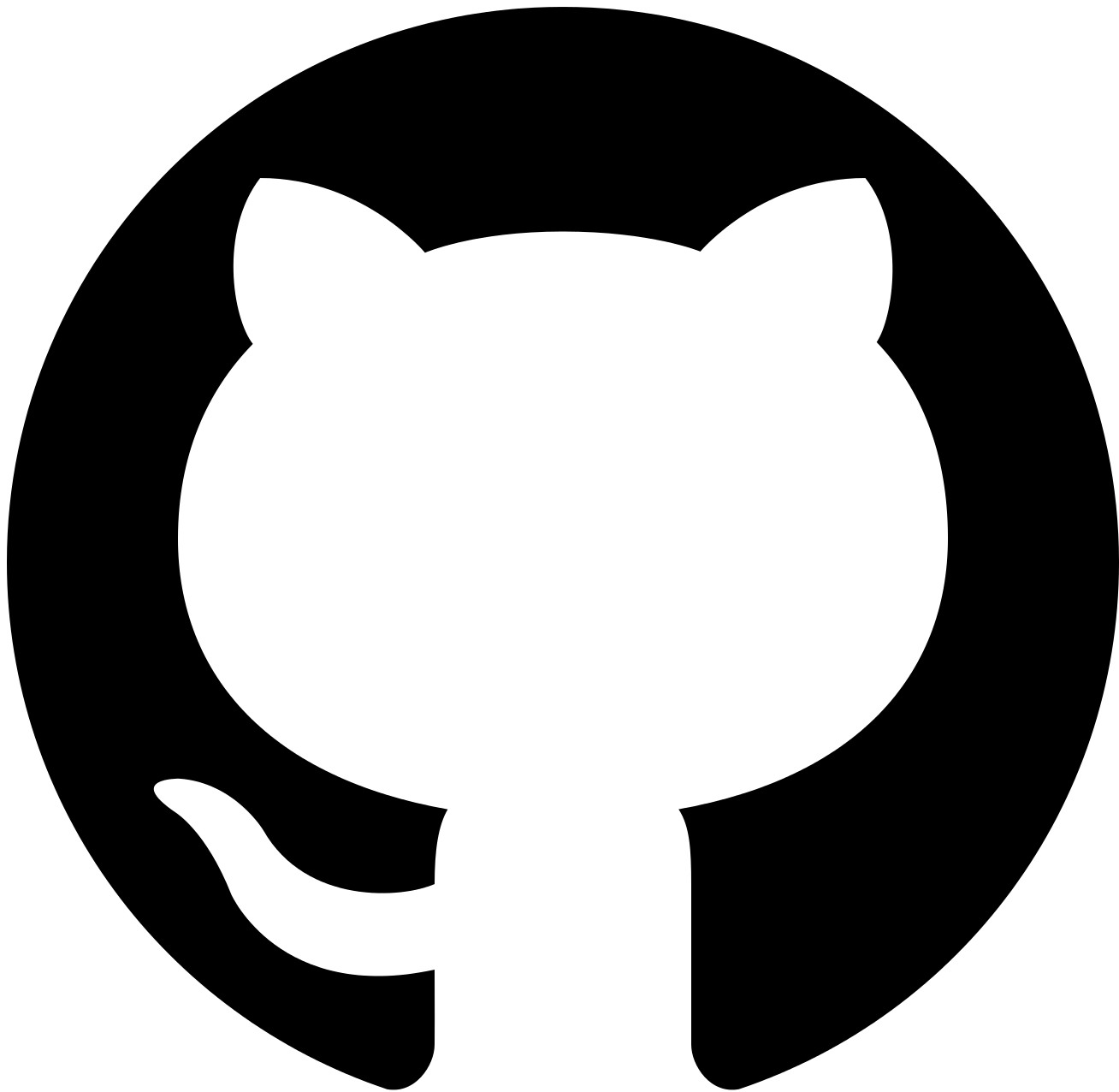}}\hspace{0.10em}\fontsize{7.5}{9}\url{https://github.com/tabular-reasoning/RUST-BENCH}
\end{abstract}

\section{Introduction}
\input{sections/intro}

\section{\AlgName Dataset}
\input{sections/rust_bench}

\section{Experiments}
\input{sections/experiments}
\input{sections/analysis}

\section{Related Work}
\input{sections/related_works}

\section{Conclusion}
\input{sections/conclusion}

\bibliography{custom}
\newpage
\appendix

\section{More Details on Dataset Generation}
\input{sections/data_generation}

\section{Implementation Details}
\input{sections/baselines}

\input{sections/error}

\end{document}

%% file: sections/intro.tex
\label{sec:intro}
Semi-structured tables containing free-form text embedded within structured fields are common across various domains~\cite{gupta2020infotabs}. Effective data analysis in science, finance, and sports requires reasoning over large, domain-specific tables that combine symbolic structure with textual context. However, existing benchmarks predominantly evaluate short, homogeneous Wikipedia-derived tables~\cite{pasupat2015compositional, chen2019tabfact}, which limits both model generalizability and robustness. Although Large Language Models (LLMs) have made tabular reasoning more accessible by allowing users to query tables directly in natural language~\cite{cheng2022binding}, systematic evaluation of their reasoning abilities over complex tables remain underexplored~\cite{chen2023large}.

% \vspace{1.0in}
\begin{figure}
  \centering
  \vspace{-5pt} % adjust vertical space if needed
  \includegraphics[width=0.99\linewidth]{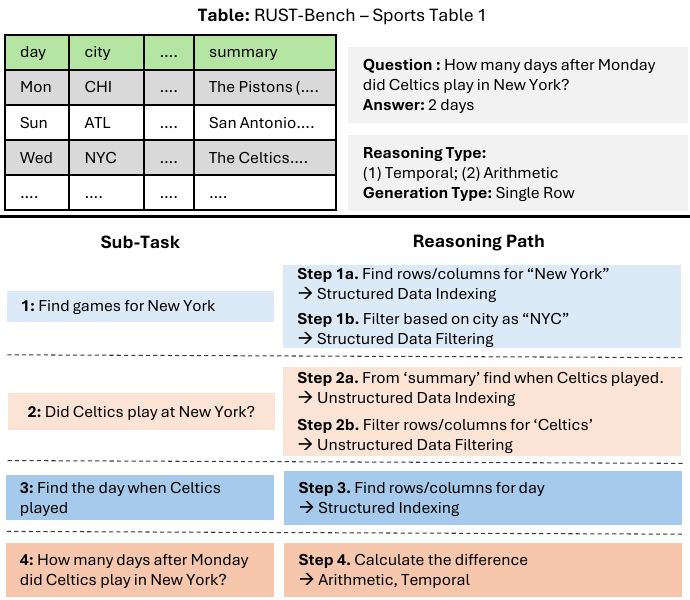}
  % \vspace{-1em}
     \caption{\textbf{Illustration of a multi-step reasoning process for a complex question grounded in a sports table from \AlgName.} The example shows that real-world tabular reasoning often demands multiple complementary reasoning skills (temporal, arithmetic, and contextual) and the coordinated use of heterogeneous evidence across long, domain-specific tables.}
  \label{fig:example_run}
  \vspace{-15pt} % adjust spacing below if needed
\end{figure}

\begin{table*}[!htbp]
    \centering
    \small
    \caption{ \small \textbf{Comparison of \AlgName with other Table QA datasets}. \AlgName contains a variety of complex question types over large, domain-specific tables containing semi-structured information. *Only the contents of the table are considered.}
    \vspace{-7pt}
    \fontsize{14}{15}\selectfont
    \resizebox{0.99\textwidth}{!}{
    \renewcommand{\arraystretch}{1.1}
    \setlength{\tabcolsep}{3pt}
    \begin{tabular}{lcccccccc}
    \toprule
    \textbf{\multirow{2}{*}{Dataset}} & \textbf{\multirow{2}{*}{Source}} & \textbf{Complex} & \textbf{Unanswerable} & \textbf{Domain} & \textbf{Semi} & \textbf{Large} & \textbf{\# Avg.} & \textbf{Context} \\
    & & \textbf{Reasoning} & \textbf{Questions} & \textbf{Specific} & \textbf{Structured} & \textbf{Tables} & \textbf{Rows} & \textbf{Length} \\
    \midrule
    WikiTQ~\cite{pasupat2015compositional}& Wikipedia~\cite{Wikipedia} & \textcolor{red}{\xmark} & \textcolor{red}{\xmark} & \textcolor{red}{\xmark} & \textcolor{red}{\xmark} & \textcolor{red}{\xmark} & 6.3 & 1133.51 \\
    TabFact~\cite{chen2019tabfact} & Wikipedia & \textcolor{red}{\xmark} & \textcolor{red}{\xmark} & \textcolor{red}{\xmark} & \textcolor{red}{\xmark} & \textcolor{red}{\xmark} & 6.2 & 586.51 \\
    Hybrid-QA~\cite{chen2020hybridqa} & Wikipedia & \textcolor{ao(english)}{\cmark} & \textcolor{red}{\xmark} & \textcolor{red}{\xmark} & \textcolor{ao(english)}{\cmark} & \textcolor{red}{\xmark} & 15.7 & 372.14$^{*}$ \\ 
    OTT-QA~\cite{chen2020open} & Wikipedia & \textcolor{ao(english)}{\cmark} & \textcolor{red}{\xmark} & \textcolor{red}{\xmark} & \textcolor{ao(english)}{\cmark} & \textcolor{red}{\xmark} & 15.7 & 372.14$^{*}$ \\ 
    CRT-QA~\cite{zhang2023crt} & Wikipedia & \textcolor{ao(english)}{\cmark} & \textcolor{ao(english)}{\cmark} & \textcolor{red}{\xmark} & \textcolor{red}{\xmark} & \textcolor{red}{\xmark} & 12.6 & 257.12 \\
    TAT-QA~\cite{zhu2021tat} & Financial Reports~\cite{Financial} & \textcolor{ao(english)}{\cmark} & \textcolor{red}{\xmark} & \textcolor{ao(english)}{\cmark} & \textcolor{ao(english)}{\cmark} & \textcolor{red}{\xmark} & 9.4 & 378.31 \\ 
    FINQA~\cite{chen2021finqa} & FinTabNet~\cite{zheng2021global} & \textcolor{ao(english)}{\cmark} & \textcolor{red}{\xmark} & \textcolor{ao(english)}{\cmark} & \textcolor{ao(english)}{\cmark} & \textcolor{red}{\xmark} & 6.4 & 687.51 \\ 
    SciTab~\cite{lu2023scitab} & SciGen~\cite{moosavi2021scigen} & \textcolor{ao(english)}{\cmark} & \textcolor{red}{\xmark} & \textcolor{ao(english)}{\cmark} & \textcolor{red}{\xmark} & \textcolor{red}{\xmark} & 7.5 & 254.53 \\ 
    \midrule
    \cellcolor{yellow!15}\textbf{\AlgName} & \cellcolor{yellow!15} NSF~\cite{NSF2024}, Sportsett~\cite{thomson2020sportsett} & \cellcolor{yellow!15} \textcolor{ao(english)}{\cmark} & \cellcolor{yellow!15} \textcolor{ao(english)}{\cmark} & \cellcolor{yellow!15} \textcolor{ao(english)}{\cmark} & \cellcolor{yellow!15} \textcolor{ao(english)}{\cmark} & \cellcolor{yellow!15} \textcolor{ao(english)}{\cmark} & \cellcolor{yellow!15} \bf 45.1 & \cellcolor{yellow!15} \bf 23040.68\\
    \bottomrule
    \bottomrule
    \end{tabular}}
\label{tab:accents}
\end{table*}
Real-world tabular reasoning introduces four major challenges for LLMs: \textbf{scale}, \textbf{multi-hop reasoning}, \textbf{heterogeneity}, and \textbf{domain specificity}. First, tables can be \textbf{long}, often spanning hundreds of rows and columns, and such long contexts are known to degrade LLM reasoning performance~\cite{liu2023lost}. Similarly, model performance deteriorates as table size grows, even when the entire table fits within the context window, since only a small fraction of rows are typically relevant to a given query~\cite{abhyankar2024h}. Second, many queries require \textbf{multi-hop reasoning}—locating relevant rows, integrating dispersed evidence, and composing it into an answer. Third, \textbf{heterogeneity} arises when tables mix structured fields with free-form text, requiring models to reason over diverse data modalities~\cite{chen2020hybridqa, zhu2021tat}. Finally, \textbf{domain specificity} introduces specialized terminology and domain-specific reasoning patterns, as seen in finance~\cite{chen2021finqa} and science~\cite{lu2023scitab}, which require specialized domain knowledge for effective inference. While existing benchmarks assess specific aspects of table reasoning, they often evaluate these challenges in isolation. \emph{The absence of benchmarks that jointly incorporate scale, heterogeneity, and domain specificity constitutes a fundamental limitation}, constraining systematic progress toward generalizable tabular reasoning models. We therefore pose the question: \emph{Can LLMs effectively reason over unstructured text embedded in long, domain-specific tables?}

To answer this, we introduce \textbf{\AlgName}, a new benchmark explicitly designed to stress-test models across \textbf{four orthogonal axes} of real-world tabular reasoning: \textit{domain specificity, table length, semi-structured information, and multi-hop reasoning}, offering a comprehensive and realistic evaluation framework. \AlgName comprises 2,031 tables primarily sourced from two domains: (a) science and (b) sports, accompanied by 7,966 carefully curated question–answer pairs. We construct the dataset using an \textit{LLM-driven hybrid symbolic–semantic generation pipeline}, that systematically constructs high-quality, multi-hop queries grounded in real-world semi-structured tables while reducing manual annotation costs. As illustrated in Figure~\ref{fig:example_run}, each question is designed to evaluate a wide spectrum of reasoning skills (including temporal, numerical, aggregation, verification, commonsense, counterfactual, and ambiguity resolution) with most requiring multi-hop reasoning that integrates information across multiple cells through both parallel and sequential inference. As shown in Table~\ref{tab:accents}, existing benchmarks primarily rely on Wikipedia, which generally involves short contexts and relatively simple reasoning. These datasets often lack domain-specific information, unanswerable queries, and large semi-structured tables, thereby limiting their capacity to appropriately reflect real-world complexity. In contrast, \AlgName introduces domain-grounded tables, expands the range of reasoning types, and substantially scales up table size (averaging 45.1 rows and roughly 23000 tokens per table). This design offers a more realistic and challenging evaluation setting for LLMs. We evaluate \AlgName using state-of-the-art proprietary and open-source LLMs, employing diverse prompting strategies and reasoning methods. Our findings expose systematic weaknesses in handling scale, heterogeneity, and reasoning composition, confirming the value of \AlgName as a challenging and diagnostic benchmark for advancing research on LLM-based table reasoning. Our main contributions are:

\noindent $\bullet$ We introduce \textbf{\AlgName}, a large-scale benchmark that jointly evaluates LLMs across four orthogonal dimensions (i.e., scale, heterogeneity, domain specificity, and complex reasoning) previously treated in isolation by existing datasets.

\noindent $\bullet$ We develop a hybrid dataset generation pipeline that leverages the complementary strengths of symbolic and semantic reasoning to construct diverse, multi-hop, domain-grounded QA pairs efficiently.

\noindent $\bullet$ Comprehensive evaluations of state-of-the-art open-source and proprietary models reveal that current LLMs struggle with large, heterogeneous tables and multi-step reasoning, exposing persistent gaps in table reasoning architectures and prompting strategies.

%% file: sections/rust_bench.tex
\label{sec:method}

\begin{figure*}[!htbp]
% \vskip 0.2in
\begin{center}
\centerline{\includegraphics[width=\textwidth]{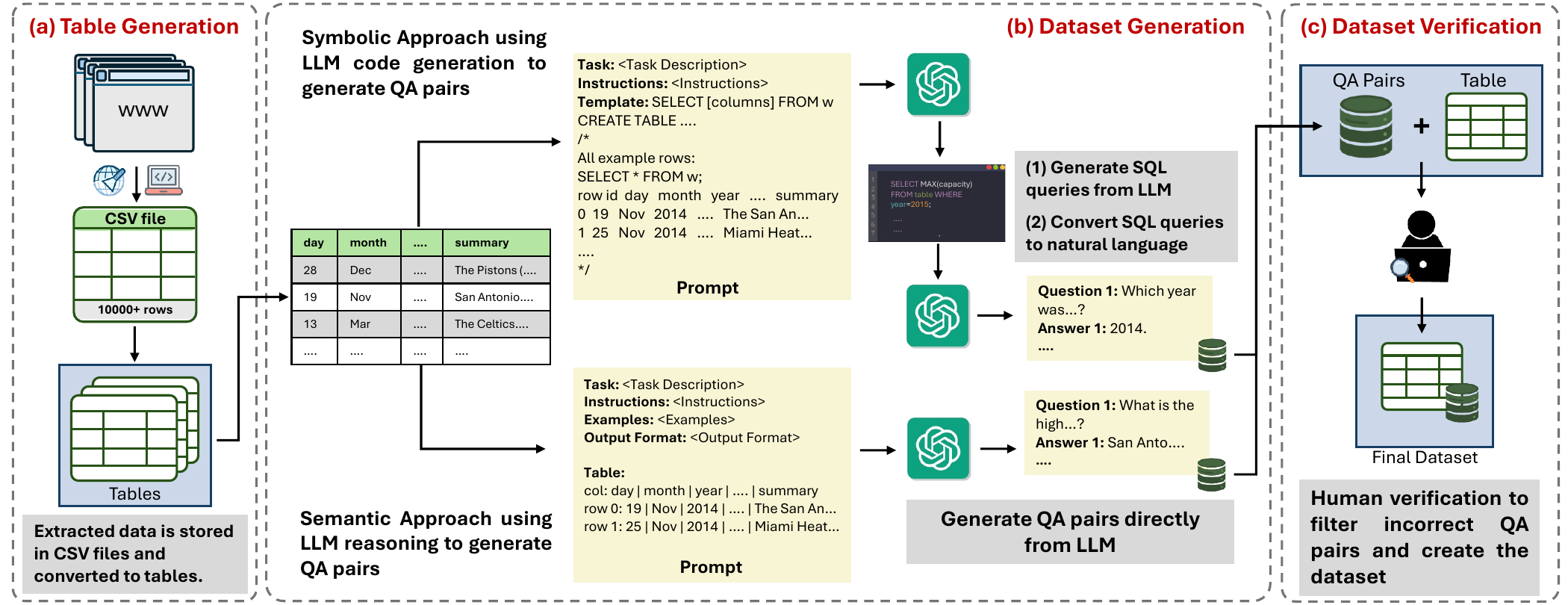}}
\caption{\small \textbf{Overview of \AlgName's dataset generation and verification pipeline.} (a) \textcolor{black}{\bf Table Generation:} Raw data are extracted from public web sources and reorganized into tables containing at least 30 rows each. (b) \textbf{Dataset Generation:}  Question–Answer pairs are created through two complementary methods: (i) a \emph{symbolic approach}, which uses SQL-like logical forms to construct schema-intensive, reasoning-heavy queries, and (ii) a \emph{semantic approach}, which employs LLMs to generate natural, inference-oriented questions from unstructured text. (c) \textbf{Dataset Verification:} All generated pairs undergo human verification to ensure factual correctness and annotation quality.}
\label{fig:main_fig}
\end{center}
\vskip -0.25in
\end{figure*}

\subsection{Task Formulation}
In table-based reasoning, each problem instance is represented as a triplet \texttt{(T, Q, A)}, where \texttt{T} denotes the tabular data, \texttt{Q} represents the associated query, and \texttt{A} signifies the anticipated response. Specifically, in the context of table-centric question-answering systems, both \texttt{Q} and \texttt{A} are in natural language. The primary objective is to derive a prediction \texttt{a} utilizing \texttt{Q} and \texttt{T}, which can be formally expressed as \texttt{a=$\pi_{\theta}$(T, Q)}, where $\pi_{\theta}$ symbolizes the predictive model.

\subsection{\AlgName Creation}

\paragraph{Table Collection.}
We curate domain-grounded tables from two high-quality sources: the NSF Grants Database~\cite{NSF2024} for science and the SportSett:Basketball dataset~\cite{thomson2020sportsett}, an enhanced version of RotoWire~\cite{wiseman2017challenges}, for sports. The raw data is cleaned and organized into domain-specific JSON tables, sampled by attributes (such as year and region) and by uniform random selection (Figure~\ref{fig:main_fig}(a)). We focus on constructing large tables with more than 30 rows, consistent with the definition in~\cite{chen2023large}. To ensure diversity and cross-domain comparability, we apply structured sampling to balance table sizes: 50\% with 30–40 rows, 40\% with 40–60, and 10\% with 60–100. This stratification balances coverage and scale across the domains, yielding a representative mixture of table sizes and schema complexities.

\paragraph{QA Generation.}
Creating high-quality QA pairs for long, domain-specific tables is particularly challenging as manual annotation is slow, costly, and prone to errors when tables span thousands of tokens. Inspired by recent LLM-based data generation methods~\cite{park2023generative, zhang2023crt, li2024planning}, we adopt in-context learning and role-playing paradigms to enable scalable and diverse dataset construction at a lower annotation cost. However, only using LLMs’ textual (semantic) reasoning is inadequate as it captures natural-language inference but fails on structural and quantitative reasoning. Conversely, symbolic reasoning methods yield precise numerical manipulation and structural consistency but lack flexibility with unstructured text~\cite{liu2023rethinking}. We therefore leverage their complementary strengths to design a hybrid symbolic–semantic pipeline~(Figure~\ref{fig:main_fig}(b)) comprising (a) a \textbf{symbolic approach}, which uses SQL-like logical forms to create schema-intensive, reasoning-heavy queries, and (b) a \textbf{semantic approach}, which generates natural, inference-oriented questions from unstructured text.

\vspace{-1.em}
\paragraph{(a) Symbolic Approach.} 

The symbolic approach exploits LLMs’ code-generation abilities to synthesize SQL queries over both structured and unstructured table components, to create questions involving numerical reasoning, aggregation, and logic. We construct a library of 75 SQL templates with placeholders (e.g., \texttt{SELECT [columns] FROM [table] WHERE [condition]}) covering diverse query patterns such as selection, aggregation, and conditional operations (Appendix~\ref{sec:symbolic}). During generation, a template is sampled and instantiated with table-specific values, providing a structural scaffold for producing valid SQL queries (Figure~\ref{fig:main_fig}(b)). For example, a template may yield \texttt{SELECT MAX(attendance) FROM RB\_Sports WHERE city==`New York'}, which is then paraphrased into a natural language question \texttt{`What is the highest attendance recorded in NYC?'} by prompting an LLM. To ensure fluency and avoid explicit SQL exposure, entity names are masked or rephrased (e.g., \texttt{New York} → \texttt{NYC}) during paraphrasing. This dual process enables coverage of multiple reasoning types, integrating structured computation with textual variation.
% \vspace{-1em}
\paragraph{(b) Semantic Approach.} The semantic component uses LLMs’ semantic reasoning to derive insights from unstructured text segments and generate diverse, inference-driven questions that go beyond surface-level lookups. However, LLMs struggle with long or complex inputs~\cite{liu2023lost}, often producing (1) overly simplistic questions and (2) repetitive patterns, especially on large tables. To mitigate these issues, we restrict inputs to either: \textbf{Single Row-Based} method for focused intra-row reasoning, or a \textbf{Multi-Row-Based} method for multi-hop reasoning across a small subset of semantically related rows. This setup reduces contextual load and encourages inference beyond simple lookups while keeping questions easily verifiable by human annotators. To further enhance diversity, we maintain a pool of in-context exemplars spanning multiple reasoning types and randomly sample from them during generation. Combined with temperature variation, this encourages broader coverage and deeper reasoning. Details of the single-row and multi-row generation processes are in Appendix \ref{sec:semantic}.

\subsection{\AlgName Validation}
Although LLMs can generate QA pairs at scale, their outputs often suffer from misalignment, limited diversity, and uneven reasoning depth~\cite{zhang2023crt}. To ensure high-quality supervision for \AlgName, we adopt a rigorous human-in-the-loop verification pipeline. This process substantially improves quality by filtering out poor generations. We first discard malformed or duplicated QA pairs and those with empty or ill-formed answers. Eight Computer Science graduate students act as annotators to review each remaining pair using a custom web interface that displays the full semi-structured table alongside its question and answer (See Appendix~\ref{sec:verif}). Annotators rate \textit{clarity, answer correctness}, and \textit{reasoning complexity} and flag uncertain or incorrect cases for secondary review. They are also instructed to ensure that the final answers are concise, self-contained, and free of redundant text to facilitate consistent automatic and human evaluation. Three expert reviewers then re-examine all pairs and consolidate the verified dataset. Low-quality or unverifiable examples are removed, while minor errors are corrected. As summarized in Table~\ref{tab:qa_verification_stats}, this process yields a curated set of high-quality QA pairs supporting multi-hop reasoning over long, heterogeneous tables.

\begin{table}[htbp]
% \vspace{-0.75em}
\centering
\caption{\small {Breakdown of QA pairs before and after human verification.}}
% \vskip 0.1in
 % \vspace{-0.5em}s
\small
\setlength{\tabcolsep}{5pt}
% \resizebox{0.5\textwidth}{!}{
\fontsize{8}{9}\selectfont
\begin{tabular}{lccccc}
\toprule
\multirow{2}{*}{\textbf{Dataset}} & \multirow{2}{*}{{\bf Category}} & \multirow{1}{*}{\textbf{Original}} & \multirow{1}{*}{\textbf{Final}} & \multirow{2}{*}{\textbf{\%~Discarded}} \\
 & &\bf~\#~QA &\bf~\#~QA & \\
\midrule
\multirow{3}{*}{\bf RB-Sports}
  & Single Row       & 2886 & 2712 & 6.0\% \\
  & Multi Row        & 1222 & 838 & 31.4\% \\
  & \multicolumn{1}{c}{Symbolic} & 1431 & 1338 & 6.5\% \\

\midrule
\multirow{3}{*}{\bf RB-Science}
  & Single Row              & 915  & 805  & 12.0\% \\
  & Multi Row               & 1516 & 1101   & 27.3\% \\
  & \multicolumn{1}{c}{Symbolic}        & 1267 & 1172   & 7.5\% \\ \bottomrule
\bottomrule
\end{tabular}
\label{tab:qa_verification_stats}
\end{table}
\vspace{-1em}

\subsection{\AlgName Statistics}
Table~\ref{tab:dataset_stats_combined} summarizes the \AlgName dataset, comprising 2,031 tables spanning RB-Sports (1,326) and RB-Science (705). Although both domains contain tables of similar length, RB-Science shows greater structural complexity, with more columns and higher token counts per table. We include 5,674 questions in RB-Sports and 2,292 in RB-Science, averaging 4.28 questions per table in RB-Sports and 3.25 in RB-Science, plus a subset of unanswerable queries. For unstructured passages, RB-Science has higher average token counts (477.62 vs. 400.58; medians 469 vs. 368) and a larger token standard deviation (149.87 vs. 114.21), while RB-Sports has slightly more sentences per passage on average. To assess annotation quality, we conducted a human-rated complexity study following~\cite{nan2022fetaqa}. Three experts rated 100 random examples on a 1–5 scale, with scores $\geq$4 indicating high-quality QA pairs. The study achieved 91.7\% inter-annotator agreement, confirming the dataset’s reliability.

% \vspace{-0.75em}
\begin{table}[!htbp]
    \centering
    \small
    \setlength{\aboverulesep}{1.5pt}
    \setlength{\belowrulesep}{1.5pt}
    \caption{\small {Summary statistics of \AlgName across RB-Sports and RB-Science.}}
    % \vspace{-0.75em}
    \setlength{\tabcolsep}{3.5pt}
    \resizebox{0.45\textwidth}{!}{
    \fontsize{7}{7}\selectfont
    \begin{tabular}{lcc}
    \toprule
     & \multicolumn{1}{c}{\bf RB-Sports} & \multicolumn{1}{c}{\bf RB-Science} \\
    \toprule
    \rowcolor[gray]{0.9}\multicolumn{3}{c}{\textbf{Tables}} \\ 
    % \midrule
    \# Tables & 1326 & 705 \\
    Avg. Rows / Table & 44.95 & 45.13  \\
    Avg. Columns / Table & 12.0 & 28.0  \\
    Avg. Tokens / Table & 18304.47 & 31948.79 \\
    \midrule
    \rowcolor[gray]{0.9}\multicolumn{3}{c}{\textbf{Questions}} \\ 
    % \midrule
    \# Questions & 5674 & 2292 \\
    Avg.~Question Length (words)  & 26.92 & 27.48  \\
    \# Questions / Table & 4.28 & 3.25 \\
    \# Unanswerable Questions & 132 & 372 \\
    \midrule
    \rowcolor[gray]{0.9}\multicolumn{3}{c}{\textbf{Unstructured Text}} \\ 
    % \midrule
    Avg Tokens / Passage & 400.58 & 477.62 \\
    Std Tokens & 114.21 & 149.87 \\
    Median Tokens & 368.00 & 469.00 \\
    Avg Sentences / Passage & 16.22 & 14.34 \\
    Std Sentences & 4.32 & 4.84 \\
    Median Sentences & 15.00 & 14.00 \\
    \midrule
    Inter-Annotator Agreement    & \multicolumn{2}{c}{\bf 91.7\%} \\
    \bottomrule
    \bottomrule
    \end{tabular}}
    \vspace{-1em}
    \label{tab:dataset_stats_combined}
\end{table}

%% file: sections/experiments.tex
\label{sec:experiments}

\begin{table*}[!htbp]
    \centering
    \caption{\small \textbf{Comparison of LLM backbones using various prompting strategies} on variants RB-Science and RB-Sports using: (a)~Exact Match (EM), (b)~BLEU, and (c)~LLM-as-a-judge (LLM-score). Higher values indicate better performance.}
    % \vspace{-0.5em}
    \small
        \setlength{\aboverulesep}{1.35pt}
    \setlength{\belowrulesep}{1.35pt}
    % \vspace{-0.75em}
    % \resizebox{\textwidth}{!}{
    \setlength{\tabcolsep}{5.5pt}
    \fontsize{9.5}{9.5}\selectfont
    \begin{tabular}{llcccccc}
    \toprule
    \bf \multirow{2}{*}{Model} & \bf \multirow{2}{*}{Strategy} & \multicolumn{3}{c}{\bf RB-Science} & \multicolumn{3}{c}{\bf RB-Sports} \\
    \cmidrule(lr){3-5} \cmidrule(lr){6-8}
     & & EM (\%) & BLEU & LLM-score (\%) & EM (\%) & BLEU & LLM-score (\%) \\
    \midrule
    \rowcolor[gray]{0.9}\multicolumn{8}{c}{\bf Large Language Models} \\
    % \midrule
    \multirow{4}{*}{GPT-4o-mini} 
        & Zero-Shot         &  36.6 & 0.293 & 40.4 & 39.8 & 0.285 & 43.1 \\
        & Few-Shot          &  37.9 & 0.296 & 36.7 & 31.3 & 0.301 & 33.9 \\
        & CoT               &  44.4 & 0.378 & 48.8 & 42.1 & 0.365 & 45.2 \\
        & PoT               &  32.8 & 0.312 & 34.5 & 30.6 & 0.285 & 33.6 \\
    \midrule
    \multirow{4}{*}{Llama-3.3-70B} 
        & Zero-Shot         &  38.8 & 0.301 & 47.1 & 39.2 & 0.311 & 44.3 \\
        & Few-Shot          &  41.7 & 0.347 & 46.4 & 46.7 & 0.350 & 48.9 \\
        & CoT               &  44.2 & 0.401 & 45.3 & 42.2 & 0.392 & 43.9 \\
        & PoT               &  27.7 & 0.299 & 30.6 & 31.1 & 0.289 & 33.0 \\
    \midrule
    \multirow{4}{*}{Gemini-2.0-Flash} 
        & Zero-Shot         &  40.7 & 0.370 & 47.3 & 38.6 & 0.345 & 45.4 \\
        & Few-Shot          &  45.9 & 0.373 & 48.8 & 41.4 & 0.340 & 43.3 \\
        & CoT               &  47.3 & 0.454 & 50.8 & 44.1 & 0.419 & 48.7 \\
        & PoT               &  18.2 & 0.225 & 23.6 & 26.3 & 0.239 & 29.1 \\
    \midrule
    \multirow{4}{*}{Mistral-Small-3.2} 
        & Zero-Shot         &  48.3 & 0.410 & 50.5 & 45.7 & 0.404 & 48.0 \\
        & Few-Shot          &  50.3 & 0.373 & 51.6 & 43.9 & 0.365 & 45.2 \\
        & CoT               &  52.6 & 0.454 & 53.1 & 51.5 & 0.446 & 51.7 \\
        & PoT               &  29.8 & 0.278 & 29.9 & 20.5 & 0.241 & 26.4 \\
    \midrule
    \rowcolor[gray]{0.9}\multicolumn{8}{c}{\bf Large Reasoning Models} \\ 
    % \midrule
    Qwen3-14B &  & 42.6 & 0.441 & 44.4 & 41.2 & 0.433 & 43.1 \\
    % \midrule
    Qwen-QwQ &  & 48.1 & 0.526 & 54.1 & 46.1 & 0.479 & 55.7 \\
    % \midrule
    Qwen-Distill-32B &  & 43.1 & 0.407 & 49.9 & 39.2 & 0.426 & 44.6 \\
    % \midrule
    Llama-Distill-70B &  & 44.6 & 0.483 & 52.4 & 40.5 & 0.455 & 50.9 \\
    \bottomrule
    \bottomrule
    \end{tabular}
    \label{tab:main_results}
    % \vspace{-0.5em}
\end{table*}

\paragraph{LLM Backbones.}
We benchmark a diverse set of state-of-the-art large language models, spanning both open-source and proprietary families, as well as reasoning-optimized variants for complex problem-solving. Specifically, we evaluate \texttt{Llama-3.3-70B-Instruct}~\cite{dubey2024llama}, \texttt{GPT-4o-mini}~\cite{openai2023gpt}, \texttt{Gemini-2.0-Flash}~\cite{team2023gemini}, and \texttt{Mistral-Small-3.2-24B-Instruct-2506}~\cite{jiang2024mixtral}. Beyond these general-purpose models, we also assess \texttt{Qwen3-14B, Qwen-32B-QwQ}~\cite{yang2025qwen3}, \texttt{Qwen-Distill-32B}, and \texttt{Llama-Distill-70B}~\cite{guo2025deepseek}, which are specialized for reasoning tasks. All models are evaluated using default hyperparameters and a fixed decoding temperature ($\tau=0.1$) for consistency across runs. Following \cite{wang2023chain}, each table is linearized into a pipe-separated format and concatenated with its query across models.

\paragraph{Baselines.}
We evaluate two baseline categories: (i)  \textbf{prompting strategies} and (ii) \textbf{table reasoning methods} developed specifically for tabular data. For prompting, we adopt four standard paradigms:
(i) \textbf{Zero-shot prompting}, where the model directly answers the table–question pair;
(ii) \textbf{Few-shot prompting}~\cite{chen2023large}, with four in-context examples;
(iii) \textbf{Chain-of-Thought (CoT)}~\cite{wei2022chain}, encouraging intermediate reasoning steps; and
(iv) \textbf{Program-of-Thought (PoT)}~\cite{chen2023program}, which incorporates executable programs as intermediate reasoning. For table reasoning methods, we use \texttt{GPT-4o-mini} and \texttt{Llama-3.3-70B} as LLM backbones to evaluate six state-of-the-art approaches: \texttt{BlendSQL}~\cite{glenn2024blendsql}, a hybrid framework embedding SQL-style reasoning within natural prompts; \texttt{Chain-of-Table}~\cite{wang2023chain}, which performs stepwise table updates for interpretable reasoning; \texttt{ProTrix}~\cite{wu2024protrix}, integrating SQL planning with compositional reasoning; \texttt{TabSQLify}~\cite{nahid2024tabsqlify}, which uses SQL to partition large tables into sub-tables for scalable inference; \texttt{TableMaster}~\cite{cao2025tablemaster}, combining textual and symbolic reasoning via adaptive table verbalization; and \texttt{NormTab}~\cite{nahid2024normtab}, normalizing table structures and values to improve symbolic interpretability. Additional implementation details are provided in Appendix~\ref{sec:baselines}.

\paragraph{Evaluation Metrics.}
For fairness and consistency, all models are evaluated under identical input and output constraints, focusing on accuracy and generation quality. Each model is instructed to produce concise, self-contained natural language answers; for SQL-based methods, query output is post-processed and verbalized in natural language for comparability. Following~\cite{pasupat2015compositional, zhang2023crt}, we report \textbf{Exact Match (EM)} as the primary metric. We further relax the evaluation with \textbf{BLEU}~\cite{papineni2002bleu} to capture $n$-gram overlap and an \textbf{LLM-as-a-Judge} (LLM-Score) evaluation using \texttt{GPT-4o-mini} to assess semantic equivalence. This combination provides complementary signals for lexical accuracy, surface fluency, and semantic faithfulness. For more details, see Appendix~\ref{sec:eval_metric}.

\subsection{Main Results}
\textcolor{black}{Table~\ref{tab:main_results} reports the performance of different LLM backbones on \AlgName using Exact Match (EM), BLEU, and LLM-score. Overall, \texttt{Qwen-QwQ} achieves the highest performance across all metrics, with an LLM-score reaching 54.1 and 55.7 for RB-Science and RB-Sports, respectively. Furthermore, it can be seen that CoT consistently outperforms Zero-Shot and Few-Shot for smaller models, highlighting the importance of explicit reasoning in this setting. In contrast, PoT exhibits the weakest performance across all models, likely due to the semi-structured nature of the data.} In Table~\ref{tab:main_baseline_multicol}, we present a comparison of table reasoning baselines on \AlgName, implemented using \texttt{GPT-4o-mini} and \texttt{Llama-3.3-70B} as the backbones. Among these, \texttt{TableMaster} achieves the best overall results, reaching 42.3\% EM on RB-Science and 43.1\% on RB-Sports. In contrast, symbolic or SQL-based methods such as \texttt{TabSQLify} and \texttt{BlendSQL} perform worse, achieving EM scores of 15.3\% and 13.6\%, respectively. These findings suggest that purely symbolic reasoning pipelines are insufficient for the flexible, context-driven inference required by \AlgName, which is consistent with our findings.

%% file: sections/analysis.tex
\label{sec:analysis}
\subsection{Impact of Table Size }
\begin{figure}[htbp]
  % \vspace{-1em}
  \centering
  \includegraphics[width=\linewidth]{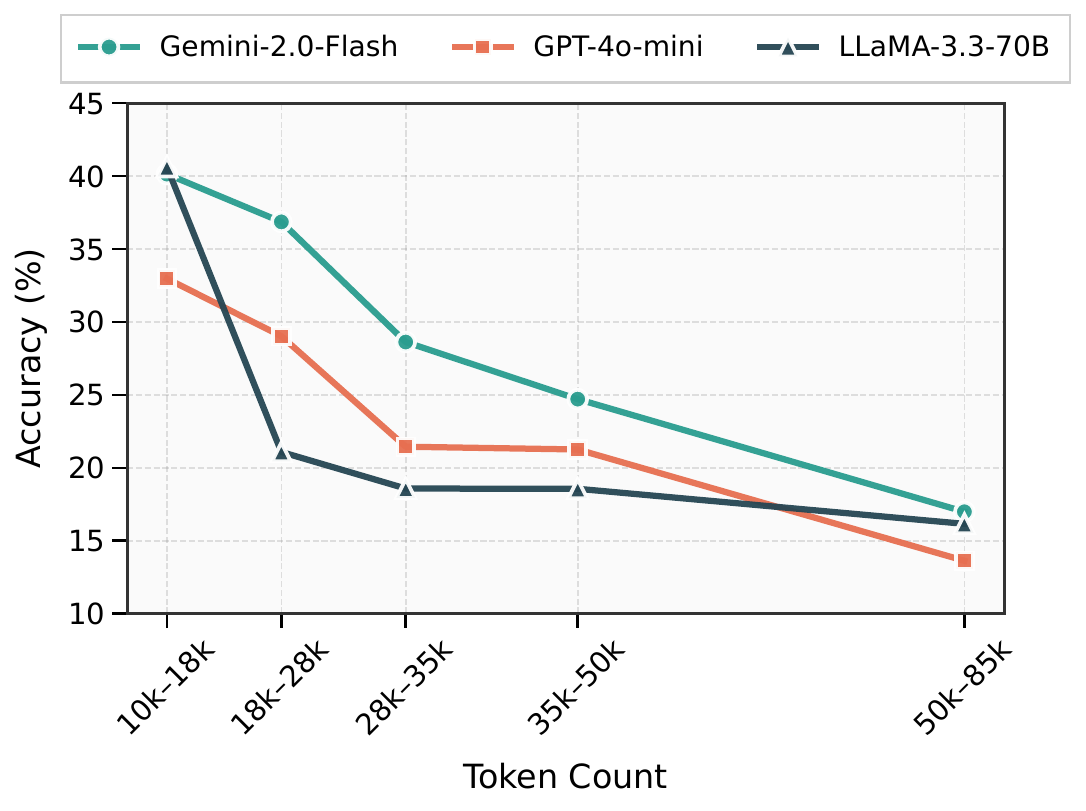}
  \vspace{-1em}
  \caption{\small {\bf Accuracy comparison of LLMs across varying token count bins}. The x-axis represents token length ranges, while the y-axis shows accuracy in percentage.}
  \vspace{-1.em}
  \label{fig:table_size}
\end{figure}

To investigate how table size affects reasoning accuracy in our setting, we analyze model performance across naturally occurring tables grouped by total token count, spanning from 10K to 85K tokens. As illustrated in Figure~\ref{fig:table_size}, \texttt{GPT-4o-mini}, \texttt{Gemini-2.0-Flash}, and \texttt{LLaMA-3.3-70B} exhibit a consistent, monotonic decline in Exact Match accuracy as table size increases, with degradation becoming particularly pronounced beyond the 35K–50K token threshold. Notably, this performance drop occurs well within the nominal context windows of modern LLMs (typically 128k+ tokens), suggesting that the bottleneck arises from reasoning and attention limitations rather than raw context length. This degradation can be attributed to LLMs' difficulty in retrieving and integrating dispersed evidence across long sequences~\cite{liu2023lost}, difficulty in locating relevant information, and increased multi-hop reasoning complexity. Unlike existing benchmarks that predominantly feature concise tables under 5000 tokens~\citep{pasupat2015compositional, chen2019tabfact}, \AlgName includes substantially longer and more heterogeneous tables where critical information is often scattered across extensive contexts. These findings highlight the need for improved query-specific data extraction mechanisms to effectively handle large-scale tabular reasoning tasks.

\begin{table*}[!htbp]
    \centering
    \caption{\small \textbf{Comparison of baselines on \AlgName using GPT-4o-mini and Llama-3.3-70B} using: (a)~Exact Match (EM), (b)~BLEU, and (c)~LLM-as-a-judge  (LLM-score), with higher values indicating better performance.}
    \small
    \setlength{\aboverulesep}{0.5pt}
    \setlength{\belowrulesep}{0.5pt}
    \vspace{-0.25em}
    \fontsize{8.5}{9}\selectfont
    \begin{tabular}{lcccccc}
    \toprule
    \multirow{2}{*}{\bf Method} & \multicolumn{3}{c}{\textbf{GPT-4o-mini}} & \multicolumn{3}{c}{\textbf{Llama-3.3-70B}} \\
    \cmidrule(lr){2-4} \cmidrule(lr){5-7}
     & EM (\%) & BLEU & LLM-score (\%) & EM (\%) & BLEU & LLM-score (\%) \\
    \midrule
TabSQLify         & 15.3 & 0.206 & 22.3 & 14.4 & 0.120 & 18.6 \\
BlendSQL          & 13.6 & 0.186  & 20.2 & 11.7 & 0.145  & 13.6 \\
ProTrix           & 32.6 & 0.319  & 33.9 & 28.3 & 0.265  & 31.5 \\
Chain-of-Table    & 30.1 & 0.247  & 35.1 & 33.2 & 0.358  & 36.9 \\
NormTab            & 33.9 & 0.338 & 36.8 & 30.9 & 0.279 & 34.9 \\
TableMaster       & 42.3 & 0.431 & 44.2 & 43.1 & 0.386 & 45.4 \\
    \bottomrule
    \bottomrule
    \end{tabular}
    % \vspace{-1.em}
    \label{tab:main_baseline_multicol}
\end{table*}

\subsection{Impact of Real-World Table Complexity} 
\begin{figure}
  \centering
  % \vspace{-2em}
  \includegraphics[width=0.99\linewidth]{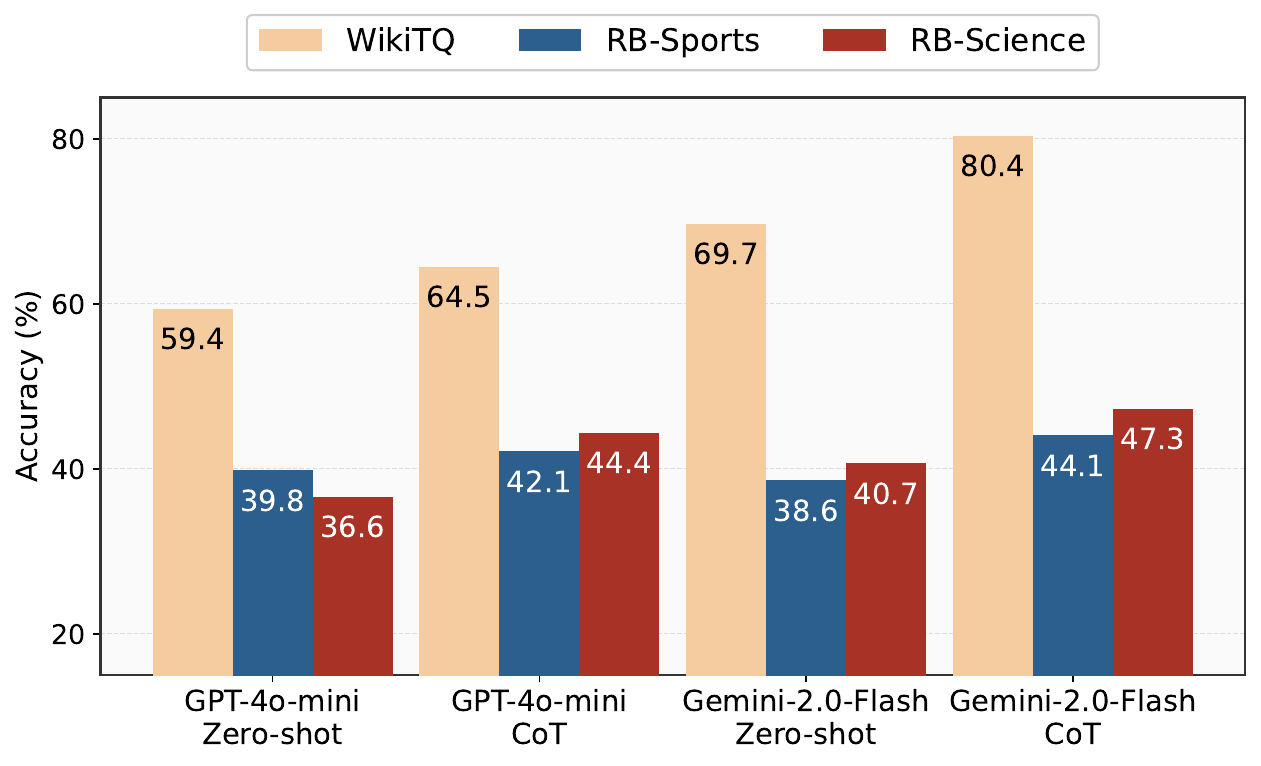}
  \vspace{-0.75em}
  \caption{\small \textbf{Performance comparison of LLM backbones on \AlgName and WikiTQ using EM accuracy.} Unlike WikiTQ, \AlgName tests LLMs with more challenging questions and tables, resulting in a reduced LLM performance.}
  \label{fig:example}
  \vspace{-1em}
% \vspace{-1em}
\end{figure}

To assess how the combination of real-world structural complexity and multi-hop reasoning affects model performance, we compare two proprietary LLMs \texttt{GPT-4o-mini} and \texttt{Gemini-2.0-Flash} across WikiTQ (a general-knowledge benchmark) and \AlgName. We evaluate both models under zero-shot and Chain-of-Thought (CoT) prompting settings. As shown in Figure~\ref{fig:example}, both models demonstrate strong performance on WikiTQ, with \texttt{GPT-4o-mini} achieving 59.4\% accuracy in zero-shot and 64.5\% with CoT, while \texttt{Gemini-2.0-Flash} reaches 69.7\% and 80.4\%, respectively. In contrast, performance on \AlgName drops sharply to roughly 20-30\% across all prompting strategies for both models. This substantial gap reveals the compounding challenges introduced by domain-specific reasoning, heterogeneous table schemas, long contexts, and multi-hop inference. Unlike WikiTQ's short, homogeneous tables dominated by direct lookup queries, \AlgName captures the full spectrum of real-world tabular reasoning, where multiple factors interact to create harder reasoning problems. Such a dramatic decline underscores the limits of current LLMs in generalizing beyond simplified benchmarks and highlights the pressing need for more robust and compositional reasoning mechanisms.

\subsection{Impact of Heterogeneous Data}
\begin{figure}[!htbp]
% \vspace{-1.em}
    \centering
    \includegraphics[width=0.99\linewidth]{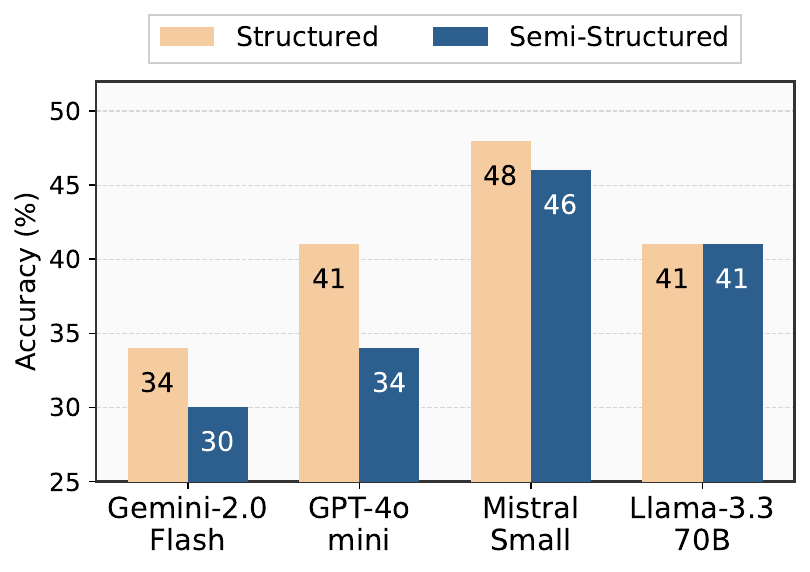}
    \vspace{-.75em}
    \caption{{\bf Performance comparison on structured and semi-structured variants} for different LLM backbones using Program-of-Thought (PoT) prompting.}
\label{fig:pot_structured_semi}
\vspace{-0.75em}
\end{figure}

While multi-hop evaluation has been extensively studied as a driver of task difficulty, the influence of data heterogeneity and structure on reasoning performance remains less explored. To investigate how the underlying data influences reasoning performance, we conduct controlled experiments on a subset of randomly sampled 100 RB-Sports tables in two settings: structured and unstructured. We convert the semi-structured tables while keeping the underlying content identical in both its variants. In the structured setting, information is normalized into explicit columns, minimizing free-form text; in the unstructured setting, each table row is verbalized into natural-language sentences and appended to the textual field, simulating highly heterogeneous inputs. We first evaluate symbolic reasoning methods, specifically Program-of-Thought (PoT) prompting, on the structured and semi-structured variants. As shown in Figure~\ref{fig:pot_structured_semi}, PoT consistently achieves higher accuracy on the structured version across all models except \texttt{Llama-3.3-70B}, which performs comparably on both. This pattern indicates that symbolic reasoning benefits from explicit schema structure and reduced textual noise, confirming its reliance on syntactic regularity. Next, we assess text-based reasoning methods using Chain-of-Thought (CoT) prompting on the unstructured and semi-structured variants. 

% \vspace{-1em}
% \vspace{-2.5em}
\begin{figure}[!htbp]
    \centering
    \includegraphics[width=0.99\linewidth]{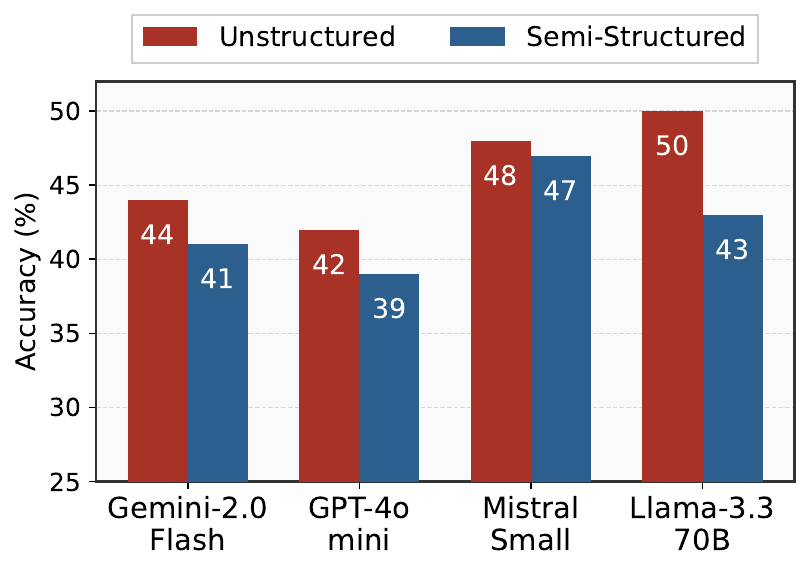}
    \vspace{-0.75em}
    \caption{{\bf Performance comparison on unstructured and semi-structured variants} for different LLM backbones using Chain-of-Thought (CoT) prompting.}
    \label{fig:cot_unstructured_semi}
     \vspace{-1.0em}
\end{figure}

Figure~\ref{fig:cot_unstructured_semi} shows that CoT yields higher accuracy on the unstructured representation, indicating that natural-language continuity facilitates stepwise reasoning when explicit structure is absent. Overall, these results show that semi-structured data presents the greatest reasoning challenge, as it combines the ambiguity of free-text with the rigidity of tabular schema, while purely structured or unstructured formats better align with the respective strengths of symbolic and semantic reasoning. We further perform an in-depth error analysis to characterize common failure modes and provide qualitative examples of reasoning diversity in \AlgName in Appendices~\ref{sec:reasoning_types}, \ref{sec:qualitative}, and~\ref{sec:error}.

%% file: sections/related_works.tex
\paragraph{General Table Reasoning.} Table reasoning tasks typically involve well-structured, short tables, often derived from Wikipedia-based sources. Datasets such as WikiTQ \cite{pasupat2015compositional}, SQA \cite{iyyer2017search}, WikiSQL \cite{zhong2017seq2sql}, and Spider \cite{yu2018spider} focus on question answering or text-to-SQL tasks that test reasoning over such tables. While WikiTQ and SQA include complex questions, WikiSQL pairs natural language questions with SQL queries, and Spider offers a large-scale, cross-domain collection with diverse databases and complex SQL. Beyond question answering, fact-verification datasets like TabFact \cite{chen2019tabfact} and Infotabs \cite{gupta2020infotabs} evaluate claim verification over Wikipedia data, while FetaQA \cite{nan2022fetaqa} targets free-form question answering requiring reasoning over entity relations. However, these datasets primarily rely on short, factual tables with limited query diversity and shallow reasoning depth.

\paragraph{Semistructured and Complex Reasoning.} Datasets such as FEVEROUS~\cite{aly2021fact}, Hybrid-QA~\cite{chen2020hybridqa}, and OTT-QA~\cite{chen2020open} extend table reasoning to open-domain contexts combining text and tables, yet still exhibit limited diversity in reasoning types and structural variation. In contrast, reasoning-focused datasets like TempTabQA~\cite{gupta2023temptabqa} and TABMWP~\cite{lu2022dynamic} emphasize specific reasoning skills like temporal and numerical reasoning, respectively, but lack semi-structured contexts. CRT-QA~\cite{zhang2023crt} covers a broader range of reasoning types but remains constrained by structured-only, open domain data. Our dataset bridges these gaps by combining domain-specific, semi-structured tables with diverse, multi-hop reasoning tasks that span both structured and unstructured modalities.

\begin{figure}[!htbp]
  \centering
  % \vspace{-7pt} % adjust vertical space if needed
  \includegraphics[width=\linewidth]{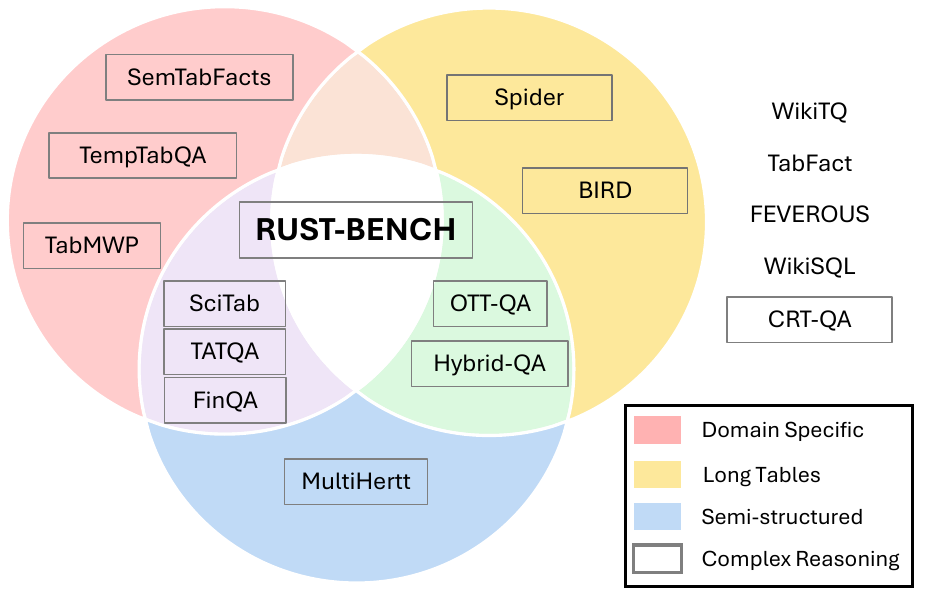}
  \vskip 0.05in
  \caption{\small \textbf{Overview of table reasoning datasets categorized by key challenges:} on (a) domain-specific, (b) long, (c) semi-structured tables, and (d) complex queries. \AlgName integrates datasets that span multiple dimensions of real-world complexity. In contrast, existing benchmarks satisfy only a subset or none of these criteria (e.g, WikiTQ, TabFact, etc.), limiting their applicability to practical, heterogeneous information systems.}
  \label{fig:illustration1}
  % adjust spacing below if needed
  % \vspace{-1.0em}
\end{figure}

\paragraph{Domain-Specific Datasets.}  Datasets tailored to specific domains typically require specialized background knowledge and retrieval mechanisms to answer domain-grounded questions. In the finance domain, FinQA~\cite{chen2021finqa}, TAT-QA~\cite{zhu2021tat}, and MultiHiertt~\cite{zhao2022multihiertt} emphasize numerical and logical reasoning, often integrating heterogeneous data sources. SemTabFacts~\cite{wang2021semeval} and SciTAB~\cite{lu2023scitab} focus on claim verification using tables from scientific articles, while SciTabQA~\cite{lu2023scitab} extends this to question answering over mixed textual and tabular evidence. Despite their domain focus, these datasets generally contain small, homogeneous tables with limited semi-structured context, thereby constraining the study of complex, multi-hop reasoning. As illustrated in Figure~\ref{fig:illustration1}, \AlgName\ differs by unifying large-scale, heterogeneous, and domain-specific tables—capturing the full spectrum of real-world reasoning challenges.

\vskip 0.05in

%% file: sections/conclusion.tex
\label{sec:conclusion}
We presented \AlgName, the first benchmark that jointly evaluates LLMs on tabular reasoning across four fundamental challenges of real-world data: scale, heterogeneity, domain specificity, and multi-hop inference. Our experiments demonstrate that even the strongest proprietary and open-source models systematically fail under these conditions, as accuracy drops sharply with increasing table length, and multi-hop reasoning over semi-structured, domain-specific tables frequently breaks down.
\AlgName provides a robust evaluation framework and a foundation for advancing research in symbolic and structured reasoning, which is an essential step toward reliable real-world deployment. Future work on \AlgName{} will emphasize \textbf{broader coverage} by adding diverse domains (e.g., healthcare, finance, climate), multilingual settings, and more complex table structures (hierarchical, nested, evolving) to better test cross-domain generalization. We will also introduce \textbf{real-world noise}, i.e., missing cells, typos, schema drift, and conflicting units—to assess robustness, calibration, and recovery under imperfect data. Finally, we will pair LLMs with tools for retrieval, schema induction, and execution, aiming for verifiable, scalable reasoning over semi-structured data.

\section*{Limitations} 
While \AlgName marks a step forward in evaluating LLMs on realistic tabular reasoning, it could further incorporate multi-table and relational reasoning, introduce training splits to support fine-tuning and adaptation, and explore richer evaluation protocols that better capture semantic correctness in complex answers. These developments can help create robust and generalizable approaches to tabular reasoning in real-world applications.

\section*{Ethics Statement}
We, the authors, affirm that our work adheres to the
highest ethical standards in research and publication. We have carefully considered and addressed various ethical issues to ensure the responsible and
fair use of computational linguistics methodologies. To facilitate reproducibility, we provide detailed information, including code, datasets (all publicly available and in compliance with their
respective ethical standards), and other relevant resources. Our claims align with the experimental results, though some stochasticity is expected with
black-box large language models, which we minimize by maintaining a fixed temperature. We provide comprehensive details on annotations, dataset
splits, models used, and prompting methods, ensuring our work can be reliably reproduced.

\section*{Acknowledgments}

This research was partially supported by the U.S. National Science Foundation (NSF) under Grant No. 2416728. We also extend our gratitude to the Complex Data Reasoning and Analysis Lab (CoRAL) at Arizona State University for providing essential computational resources, mentorship, and a collaborative research environment that greatly contributed to the progress of this work. We sincerely thank Beenaa Salian and Preethi Suresh for their assistance in data annotation, verification, and code implementation, which played a key role in ensuring the accuracy and reliability of our results. Finally, we appreciate the thoughtful and constructive feedback provided by the reviewers, which helped strengthen the quality and presentation of this research.

%% file: sections/data_generation.tex
\label{sec:data_gen}
\subsection{Symbolic Approach}
\begin{figure}[!htbp]
  % \vspace{-2em}
  \centering
  \includegraphics[width=\linewidth]{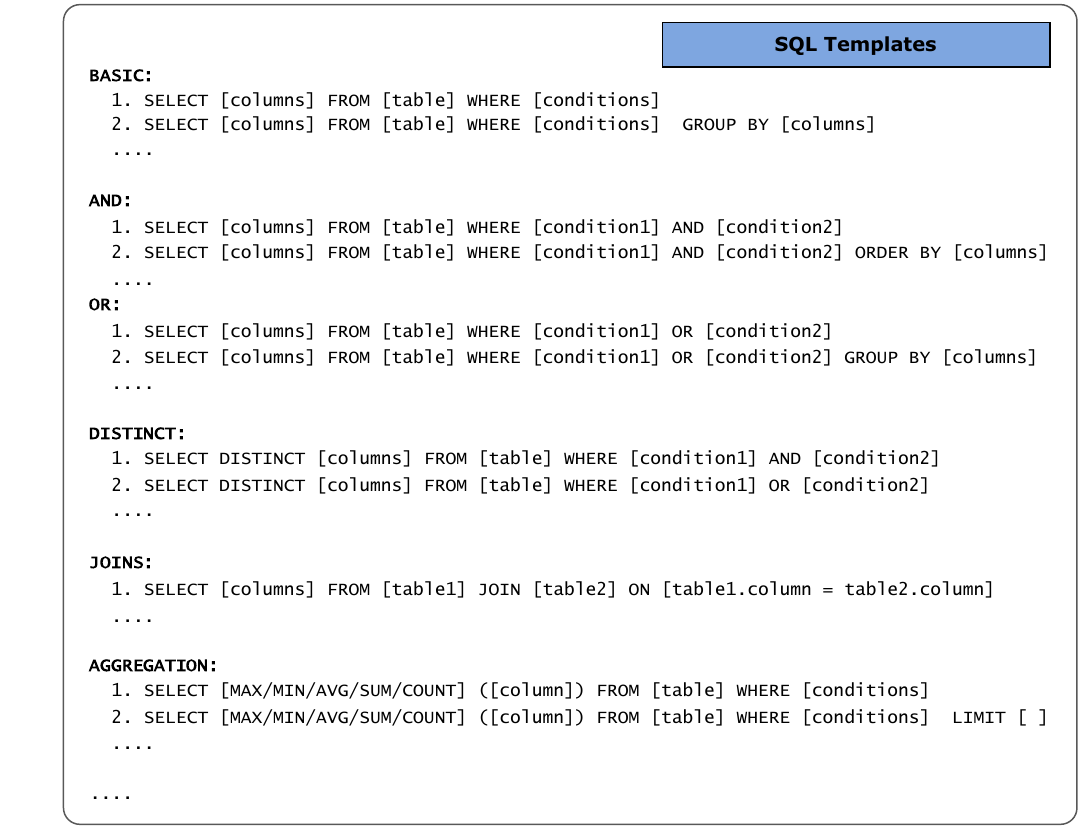}
  % \vspace{-1em}
  \caption{Example of SQL templates used for QA generation.}
  % \vspace{-.5em}
  \label{fig:templates}
\end{figure}

\begin{figure*}[!htbp]
\centering
\begin{subfigure}[b]{0.8\textwidth}
    \centering
    \includegraphics[width=\textwidth]{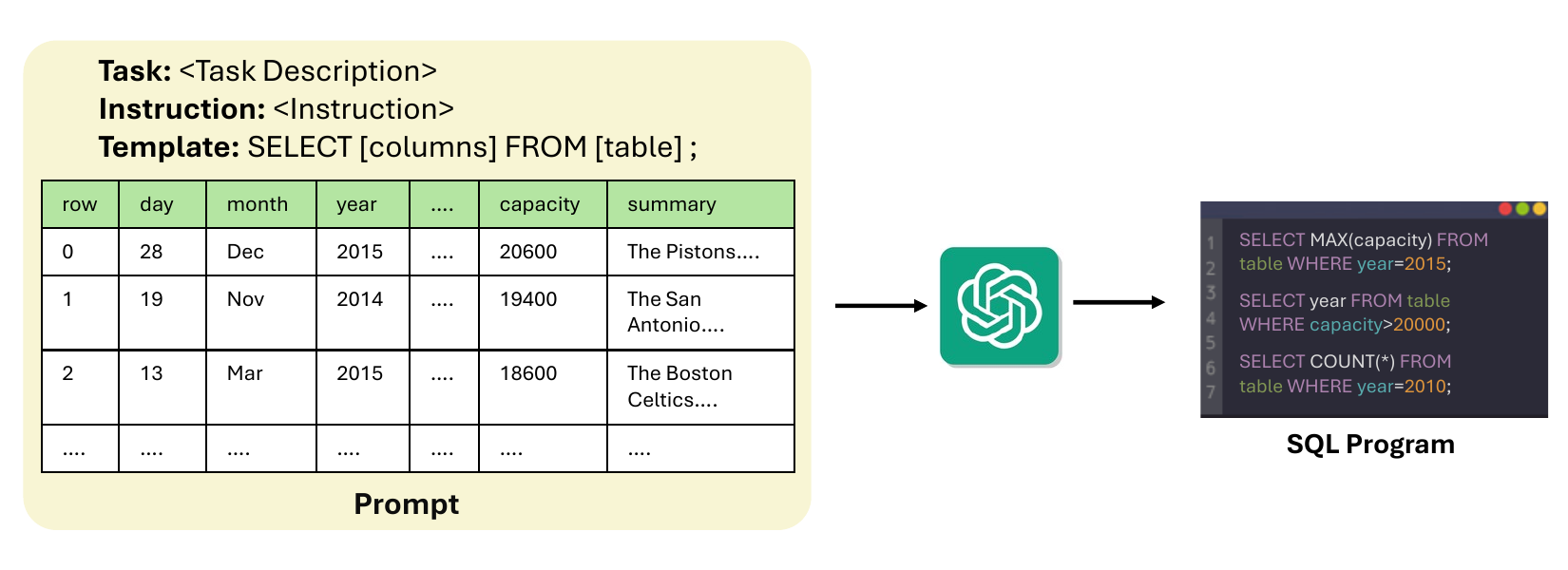}
    \vspace{-1.25em}
    \caption{\small Using code generation capabilities of LLMs to generate SQL queries.}
    \label{fig:semantic_fig_top}
\end{subfigure}

% \vspace{0.5cm}

\begin{subfigure}[b]{0.8\textwidth}
    \centering
    \includegraphics[width=\textwidth]{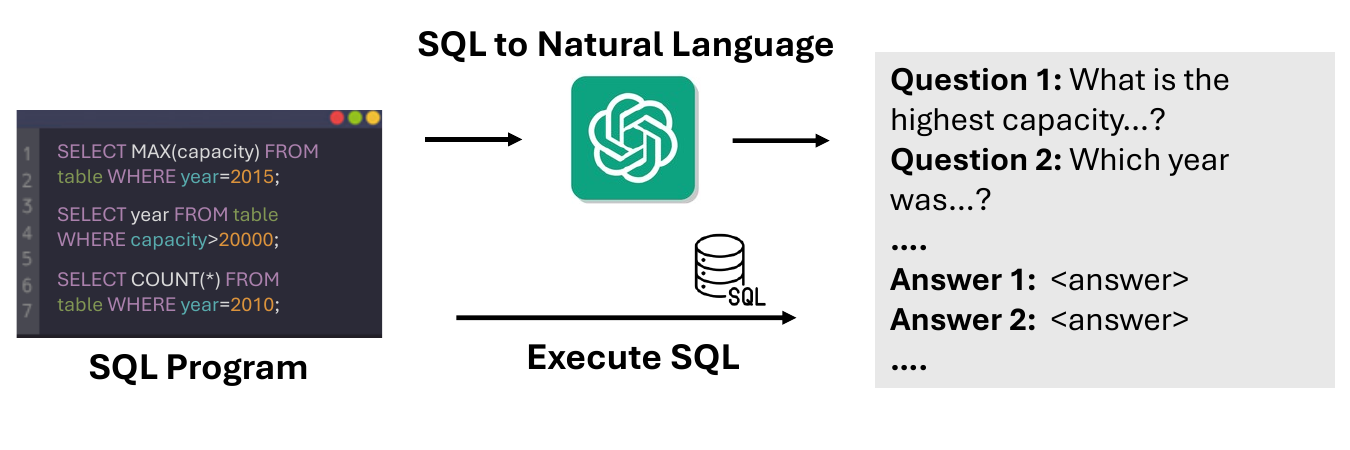}
    \vspace{-1.5em}
    \caption{\small Converting the SQL queries to natural language question-answer pairs.}
    \label{fig:symbolic_fig_1}
\end{subfigure}

\caption{\small \textbf{QA pair generation using symbolic approach.} We leverage LLMs' code generation capabilities to generate SQL queries, which are then converted to natural language questions and answers by executing the SQL queries on the table data.}
\label{fig:symbolic_fig}
\end{figure*}

To enable QA pair generation using the symbolic approach, we curate a diverse collection of approximately 75 SQL query templates. These templates are designed to cover a broad spectrum of SQL constructs, including basic \texttt{SELECT} statement, conditional logic (\texttt{AND, OR}), aggregation (\texttt{MAX, SUM, etc.}), sorting (\texttt{ORDER BY}), grouping (\texttt{GROUP BY}), and joins. As shown in Figure~\ref{fig:templates}, each template includes placeholder tokens for table names, columns, and filter conditions, allowing for broad applicability across different schemas. To instantiate these templates, we adopt a prompt-based generation approach leveraging large language models (LLMs). Specifically, we sample a template at random and prompt the LLM with task instructions and in-context exemplars to replace the template placeholders using schema-specific information derived from a target semi-structured table. This results in a fully instantiated SQL query tailored to the table (Figure~\ref{fig:symbolic_fig}). The generated SQL is then executed on the underlying table to obtain the corresponding answer. In a subsequent step, we prompt the LLM with the SQL query and its result to generate a natural language question that semantically aligns with the query logic but obscures the clauses. The final output is a question-answer pair, where the answer is grounded in the execution result of the SQL, and the question is a fluent natural language version reflecting the underlying semantics. This pipeline supports scalable QA dataset generation grounded in executable symbolic programs, enabling evaluation of models on structured reasoning tasks.

\label{sec:symbolic}

\subsection{Semantic Approach}
\label{sec:semantic}
\begin{figure*}[!htbp]
\centering
\begin{subfigure}[b]{\textwidth}
    \centering
    \includegraphics[width=\textwidth]{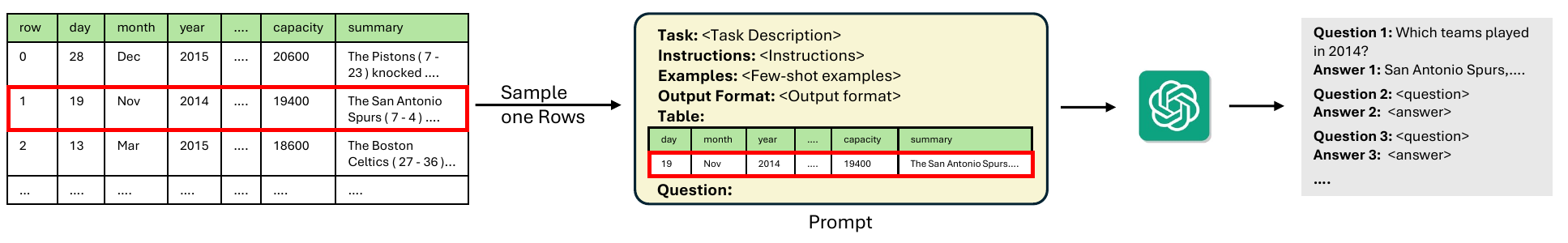}
    % \caption{\small Single-Row Approach}
    \label{fig:semantic_fig_top}
\end{subfigure}

% \vspace{-.5cm}

\begin{subfigure}[b]{\textwidth}
    \centering
    \includegraphics[width=\textwidth]{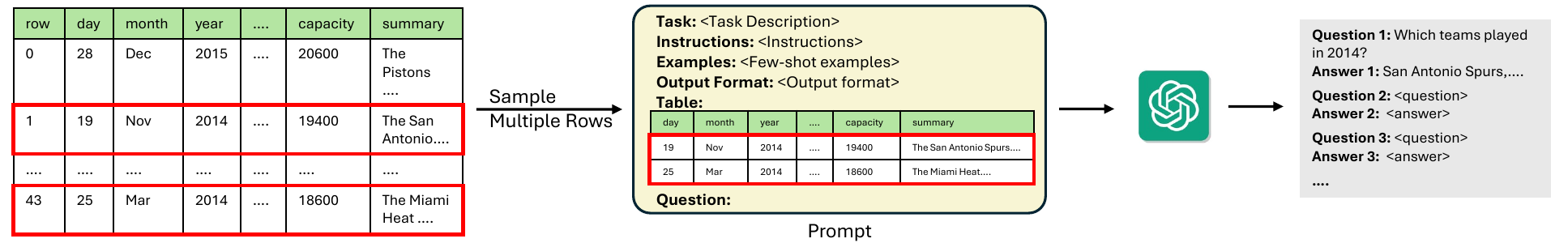}
    % \caption{\small Multi-Row Approach}
    \label{fig:semantic_fig_bottom}
\end{subfigure}
% \vspace{-3em}
\caption{\small \textbf{QA pair generation using semantic approach:} (a)~Single-Row Approach (top); (b)~Multi-Row Approach (bottom), which forms questions on a subset of the table.}
\label{fig:semantic_fig}
\end{figure*}

As outlined in Section~\ref{sec:method}, we employ two prompting strategies: Single Row-Based and Multi-Row-Based to improve the quality, diversity, and verifiability of LLM-generated questions over large tabular data. Figure~\ref{fig:semantic_fig} illustrates both approaches. In the Single Row-Based method, we randomly sample one row from the table and use it as the entire input context. This localization helps the LLM focus on intra-row reasoning, such as retrieving or interpreting structured and unstructured cell content. It also simplifies verification, as each question-answer (QA) pair depends on a well-defined and constrained context. In contrast, the Multi-Row Based method is designed to enable multi-row reasoning by selecting a subset of rows that are semantically connected via a shared entity in a specific column. By narrowing the input to only a few rows, these strategies, as shown in Figure~\ref{fig:semantic_fig}~(bottom), help overcome LLM limitations with long inputs by explicitly controlling context size and composition. They allow generating QA pairs that are diverse in type, grounded in the table content, and more easily verifiable.

% \vspace{-1.5em}

% \vskip -0.1in

\subsection{More Details on Data Validation}
\label{sec:verif}

\begin{figure*}
  \vspace{-0.75em}
  \centering
  \includegraphics[width=\linewidth]{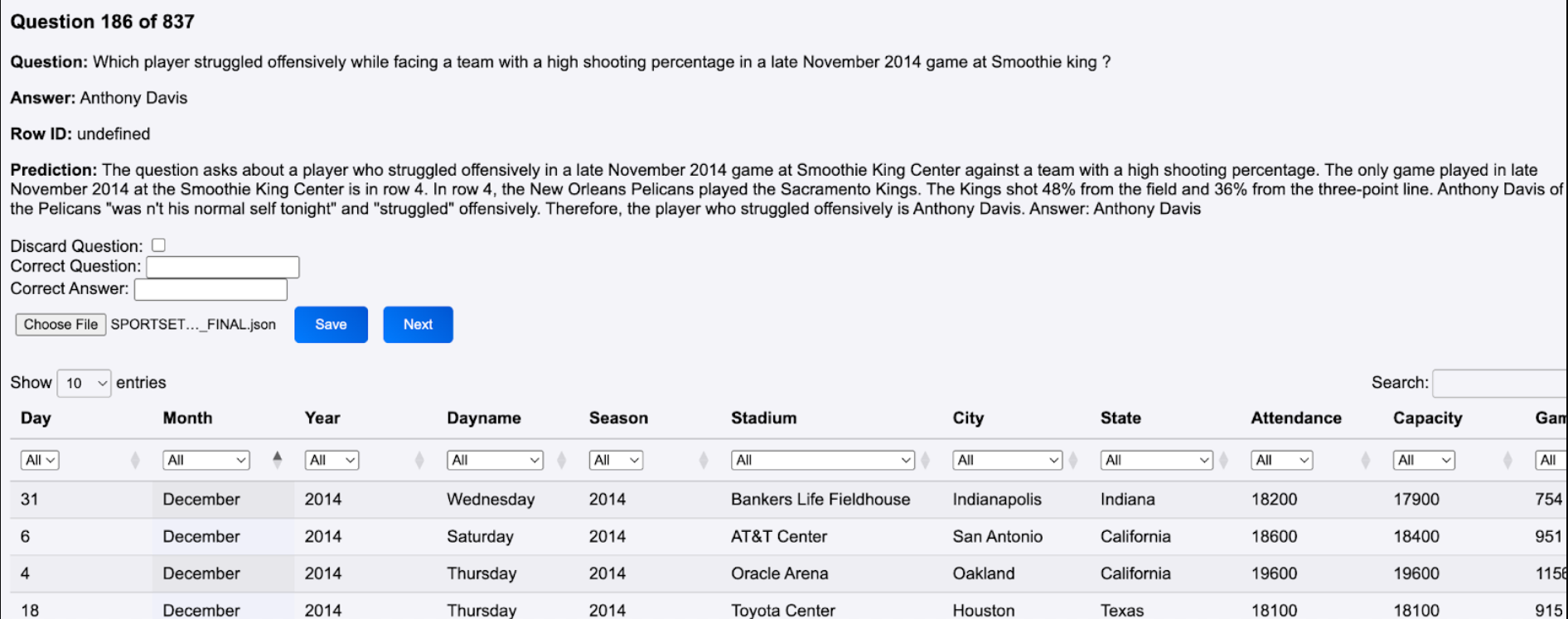}
  \vspace{-1.5em}
  \caption{Annotation Platform - User Interface.}
  \vspace{-.75em}
  \label{fig:annotation_ui}
\end{figure*}

Figure~\ref{fig:annotation_ui} illustrates the custom verification interface used during the human-in-the-loop annotation process. Each screen presents a question, its predicted answer, and a detailed explanation generated by the model, alongside an interactive table view displaying the relevant semi-structured data. Annotators could validate the question-answer pair using tools such as column-specific filters, row-level sorting, and a search bar to locate supporting evidence quickly. The interface also includes input fields for correcting errors and a checkbox for discarding invalid questions. This setup ensured that annotators had full contextual access while verifying QA pairs, improving both accuracy and efficiency. After one round of annotations, the samples were further verified by expert verifiers to ensure high-quality question-answer pairs. The entire process was conducted by annotators and reviewed by graduate students in Computer Science.

% \begin{figure}
%     \centering
%     \includegraphics[width=0.5\linewidth]{images/ui.png}
%     \caption{Caption}
%     \label{fig:annotation_ui}
% \end{figure}

%% file: sections/baselines.tex
In this section, we describe the prompting strategies, evaluation metrics, and LLM-based table reasoning baselines used in our study, along with their implementation details.

\subsection{Prompting Techniques}
\label{sec:llms}
\begin{figure*}
    \centering
    \includegraphics[width=0.8\linewidth]{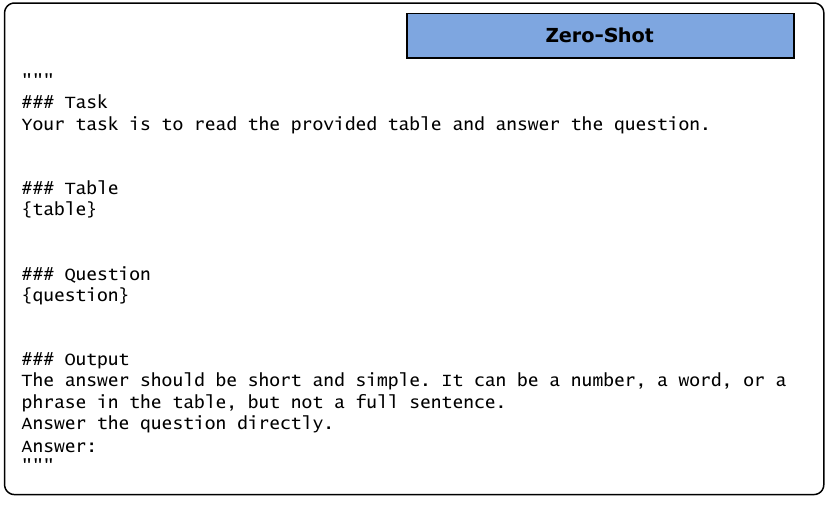}
    \caption{Prompt for Direct prompting.}
    \label{fig:zero-shot}
\end{figure*}

\begin{figure*}
    \centering
    \includegraphics[width=0.8\linewidth]{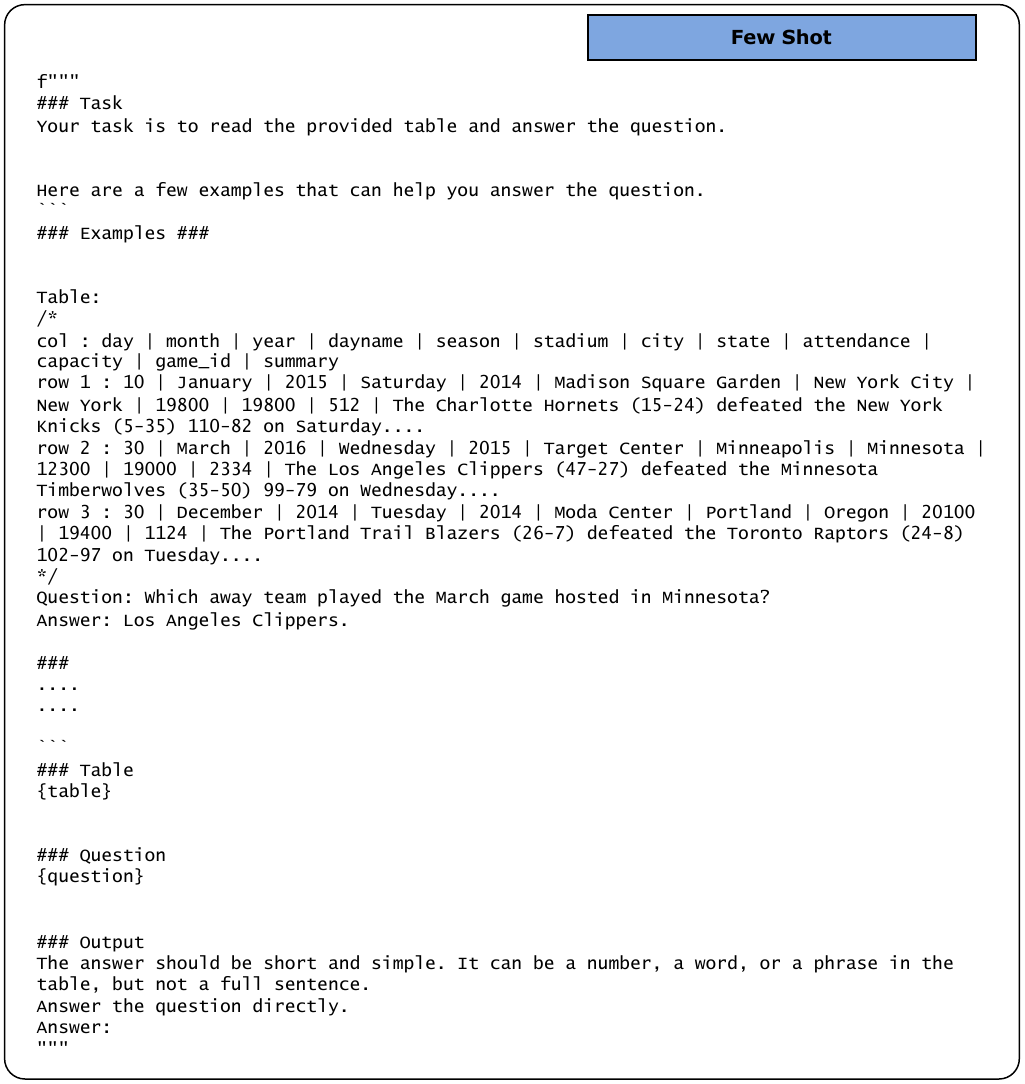}
    \caption{Prompt for Few Shot reasoning.}
    \label{fig:few-shot}
\end{figure*}

\begin{figure*}
    \centering
    \includegraphics[width=0.8\linewidth]{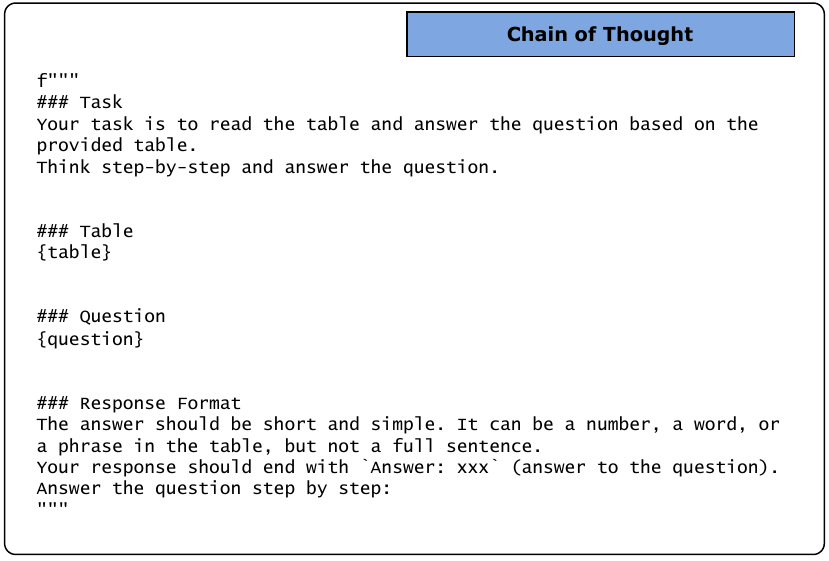}
    \caption{Chain-of-Thought reasoning prompt.}
    \label{fig:cot}
\end{figure*}

\begin{figure*}
    \centering
    \includegraphics[width=0.8\linewidth]{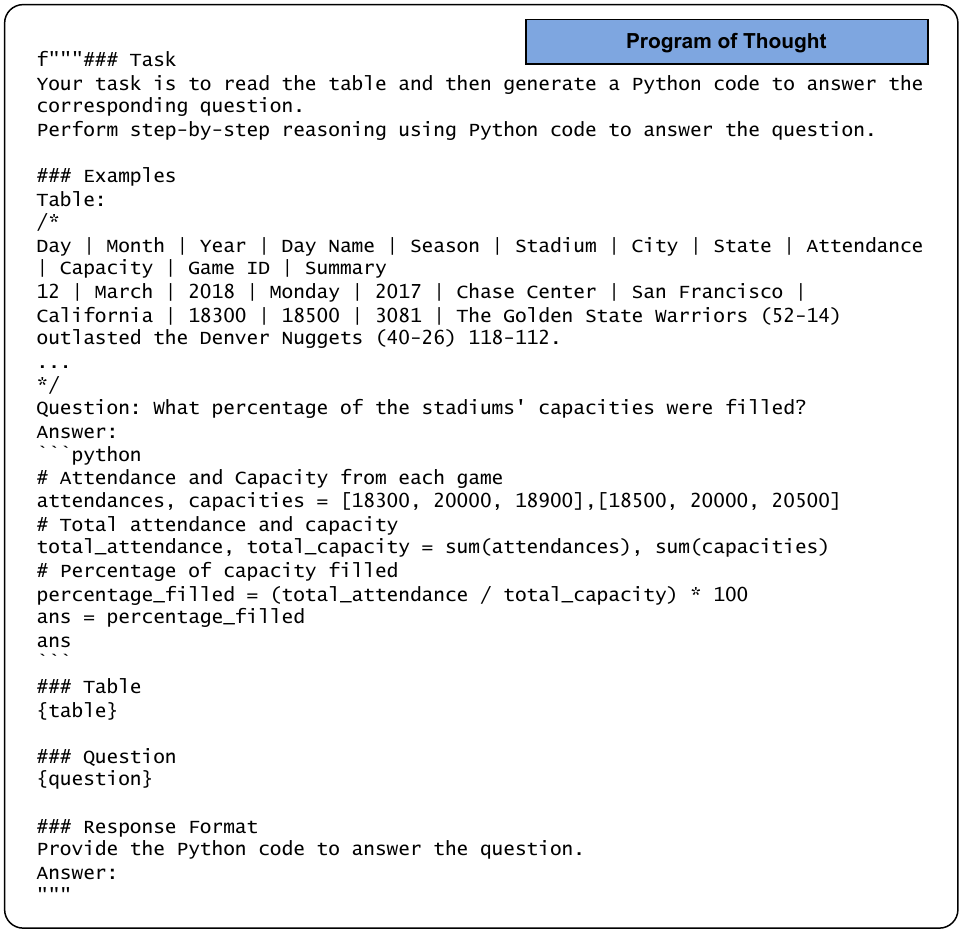}
    \caption{Program-of-Thought reasoning prompt.}
    \label{fig:pot}
\end{figure*}

We implement four reasoning techniques to use LLMs to perform tabular reasoning. Figures~\ref{fig:zero-shot},~\ref{fig:few-shot}, ~\ref{fig:cot}, and~\ref{fig:pot} highlight the direct prompting (zero-shot), few-shot, chain-of-thought (CoT) and program-of-thought (PoT) prompts for the LLMs respectively.

\subsection{Evaluation Metrics}
\label{sec:eval_metric}

% % \vspace{-.45em}

\paragraph{Exact Match (EM).} Following WikiTQ~\cite{pasupat2015compositional}, we implement exact match (EM) as the metric for evaluating model performance. EM assigns a score of 1 if the predicted answer is exactly the same as the gold answer, and 0 if otherwise. The final EM accuracy is calculated by adding the individual exact match scores divided by the total number of samples in the set. However, despite ignoring regex, punctuations, and case-sensitivity, EM penalizes semantically correct generations that do not exactly match the ground truth. It becomes increasingly challenging to evaluate longer answers that contain short phrases or multiple entities as the answer. We thus explore more relaxed metrics that do not penalize semantically correct generations.
% \vspace{-.45em}

\paragraph{BLEU Score.} BLEU score~\cite{papineni2002bleu} is a metric used in machine translation to compare the quality of machine-translated text with a set of reference translations. It measures the n-gram overlap between the reference text and the prediction, assigning a score of 0-1 depending on the amount of overlap. Despite being better than EM at longer phrases, the BLEU score measures the word overlap, missing out on the semantic relevance between the prediction and the reference.
% \vspace{-.45em}

\begin{figure*}[htbp]
    \centering
    \includegraphics[width=0.7\linewidth]{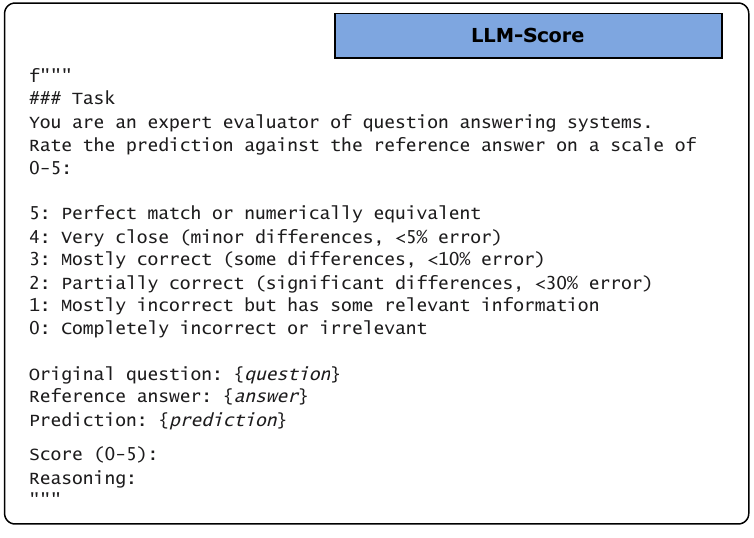}
    \vspace{-1.em}
    \caption{Prompt for using LLM-as-a-judge to output LLM-score.}
    \label{fig:llm-score}
\end{figure*}

\paragraph{LLM-score.} To correctly measure the generation quality and take the semantic similarity between the outputs and the predictions, we use an LLM as a judge to evaluate and score the generated outputs. As illustrated in Figure~\ref{fig:llm-score}, the LLM is tasked to assign a score on a scale of 0-5 based on the correctness of the prediction. With a score of 4 representing less than 5\% error between the ground truth and the prediction, the final accuracy is calculated by summing the total number of samples reporting a score of 4 or more, divided by the total samples. This enables us to gauge the answers semantically and return a better metric to evaluate the answers semantically.

\subsection{Baselines}
\label{sec:baselines}
% \vspace{-.5em}

\paragraph{BlendSQL}~\cite{glenn2024blendsql}  is a unified dialect that integrates SQL logic with large language model (LLM) reasoning across semi-structured data. It serves as a superset of SQLite, enabling complex hybrid question answering tasks involving multi-hop reasoning. The implementation utilizes the open-source repository {\texttt{blendsql}\footnote{\url{https://github.com/parkervg/blendsql}}}, with dataset-specific in-context examples and default parameters.
\vspace{-.25em}

\paragraph{Chain-of-Table}~\cite{wang2023chain} is a prompting framework that extends Chain-of-Thought by incorporating tabular data explicitly in the reasoning chain. It guides LLMs using in-context learning to iteratively generate operations and update the table to represent a tabular reasoning chain. The implementation follows the official GitHub repository \texttt{chain-of-table}\footnote{\url{https://github.com/google-research/chain-of-table}} with the in-context examples tailored to our dataset.

\vspace{-.25em}

\paragraph{ProTrix}~\cite{wu2024protrix} introduces a Plan-then-Reason framework that plans the reasoning path using the query and context, then assigns each step to either textual or program-based reasoning to arrive at the final answer. We modify their official repository\footnote{\url{https://github.com/WilliamZR/ProTrix}} in-context examples to suit \AlgName and use their default hyperparameters.
\vspace{-.25em}

\paragraph{TabSQLify}~\cite{nahid2024tabsqlify}  is a semantic parsing-based method that translates natural language questions into executable SQL queries over structured tables. It leverages text-to-SQL generation to decompose tables into smaller, relevant sub-tables containing only essential information for answering questions or verifying statements. We utilize \texttt{tabsqlify}\footnote{\url{https://github.com/mahadi-nahid/TabSQLify}} with updated in-context examples for inference.
\vspace{-.25em}

\paragraph{TableMaster}~\cite{cao2025tablemaster} is a unified framework that combines multiple techniques for table reasoning. The method first retrieves relevant table content and enriches it with semantic verbalizations, and employs adaptive reasoning to flexibly choose between textual and symbolic reasoning depending on each query. We adopt the official repository \texttt{TableMaster}\footnote{\url{https://github.com/zzlang-c/TableMaster}}, retaining their default hyperparameters for fair comparison.

\vspace{-.25em}

\paragraph{NormTab}~\cite{nahid2024normtab} focuses on improving symbolic interpretability by normalizing table structures and values prior to reasoning. It standardizes heterogeneous column names and formats, reducing schema variance and enabling more consistent SQL-based reasoning across diverse tables. We utilize the public \texttt{normtab}\footnote{\url{https://github.com/mahadi-nahid/NormTab}} repository, following default parameters and adapting the prompts to our dataset.

%% file: sections/error.tex
\section{Reasoning Diversity in \AlgName}
\label{sec:reasoning_types}
\paragraph{Distribution of Question Types.}
To understand the reasoning diversity in \AlgName, we adopt and extend the taxonomy proposed in CRT-QA~\cite{zhang2023crt}, which builds on the BIG-bench framework~\cite{srivastava2022beyond}. As shown in Table~\ref{tab:reasoning-types}, our annotation covers a broad spectrum of reasoning types—from high-frequency operations such as filtering and temporal reasoning to more complex forms including multi-hop, implicit, and counterfactual reasoning. This diversity underscores the layered cognitive demands required for real-world table understanding. Filtering and temporal reasoning are the most common types, reflecting the frequent need to locate relevant records and interpret time-dependent relationships. However, a significant proportion of questions also require multi-hop reasoning (26.18\%), numerical computation (26.83\%), and logical composition (27.85\%), highlighting the dataset’s emphasis on compositional and quantitative reasoning. Although rarer, counterfactual, commonsense, and causal reasoning further test model generalization beyond surface-level retrieval.

\begin{table}[!h]
% \vspace{-0.5em}
\centering
\caption{Distribution of reasoning types. Categories are non-exclusive; percentages may not sum to 100\%.}
% \vspace{-0.5em}
\resizebox{0.4\textwidth}{!}{
\fontsize{5}{5}\selectfont
\begin{tabular}{ll}
\toprule
\textbf{Reasoning Type} & \textbf{Percentage (\%)} \\
\midrule
 Filtering~/~Selection & 75.89 \\
Temporal Reasoning & 39.33 \\
 Logical Reasoning & 27.85 \\
Numerical & 26.83 \\
 Multi-hop Reasoning & 26.18 \\
Aggregation & 23.97 \\
 Comparison & 17.43 \\
Implicit Reasoning & 11.36 \\
 Unanswerable & 6.83 \\
Sorting~/~Ranking & 5.47 \\
 Causal Reasoning & 5.20 \\
Commonsense Reasoning & 0.44 \\
 Spatial Reasoning & 0.24 \\
Counterfactual~/~Negative & 0.19 \\
\bottomrule
\bottomrule
\end{tabular}}
\label{tab:reasoning-types}
\end{table}

\begin{figure}[!htbp]
  \centering
  % \vspace{-1em}
  \includegraphics[width=\linewidth]{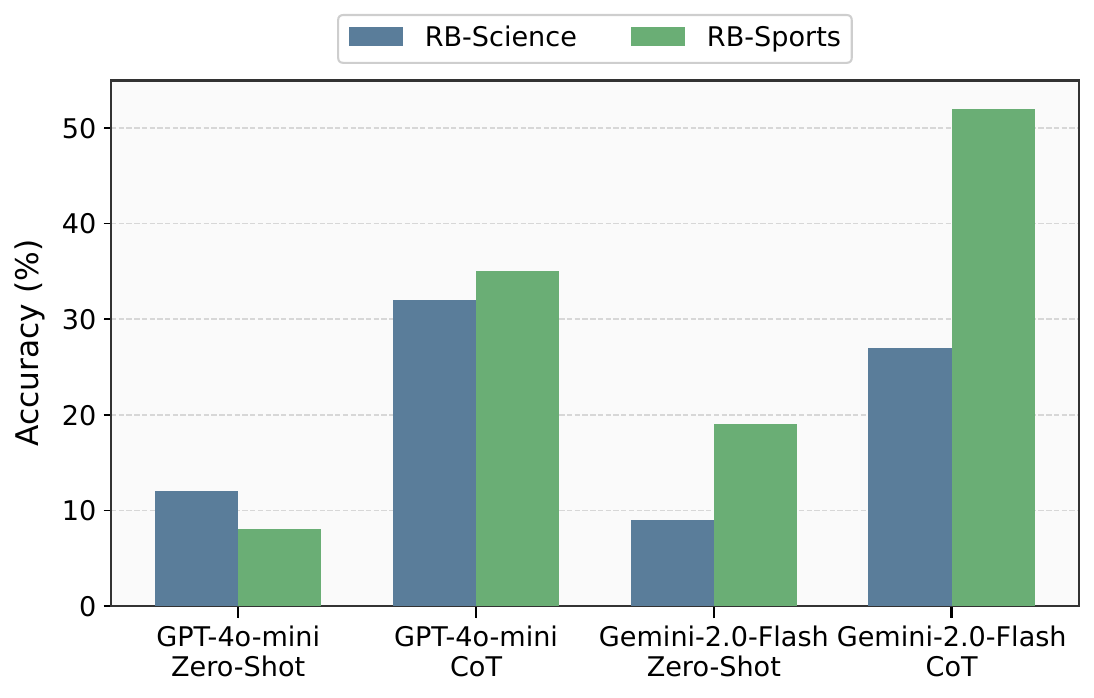}
  \caption{\small \textbf{Accuracy of GPT-4o-mini and Gemini-2.0-Flash} models on RB-Science and RB-Sports datasets, evaluated on questions that include unanswerable/ambiguous cases.}
  \label{fig:unanswerable}
\vspace{-1em}
\end{figure}

\paragraph{Unanswerable Questions.} 
In practical table reasoning, not all queries are grounded in the available data. Distinguishing answerable from unanswerable questions is therefore crucial for reliable model deployment in domains such as finance and science. To evaluate this capability, \AlgName incorporates explicitly unanswerable questions following~\cite{zhang2023crt}—queries that cannot be resolved using the table content alone. Examples include those that require external knowledge or contain logical contradictions. A model is considered correct only if it abstains by responding with phrases such as “cannot answer” or “not enough information.” We manually verify outputs to measure accuracy. As shown in Figure~\ref{fig:unanswerable}, models struggle considerably with this task: even under Chain-of-Thought prompting, \texttt{Gemini-2.0-Flash} achieves only 52.27\% accuracy in RB-Sports and 26.97\% in RB-Science, indicating the persistent challenge of reliable unanswerable detection in table QA.

\section{Qualitative Analysis}
\label{sec:qualitative}
\begin{figure*}
  \vspace{-0.75em}
  \centering
  \includegraphics[width=\linewidth]{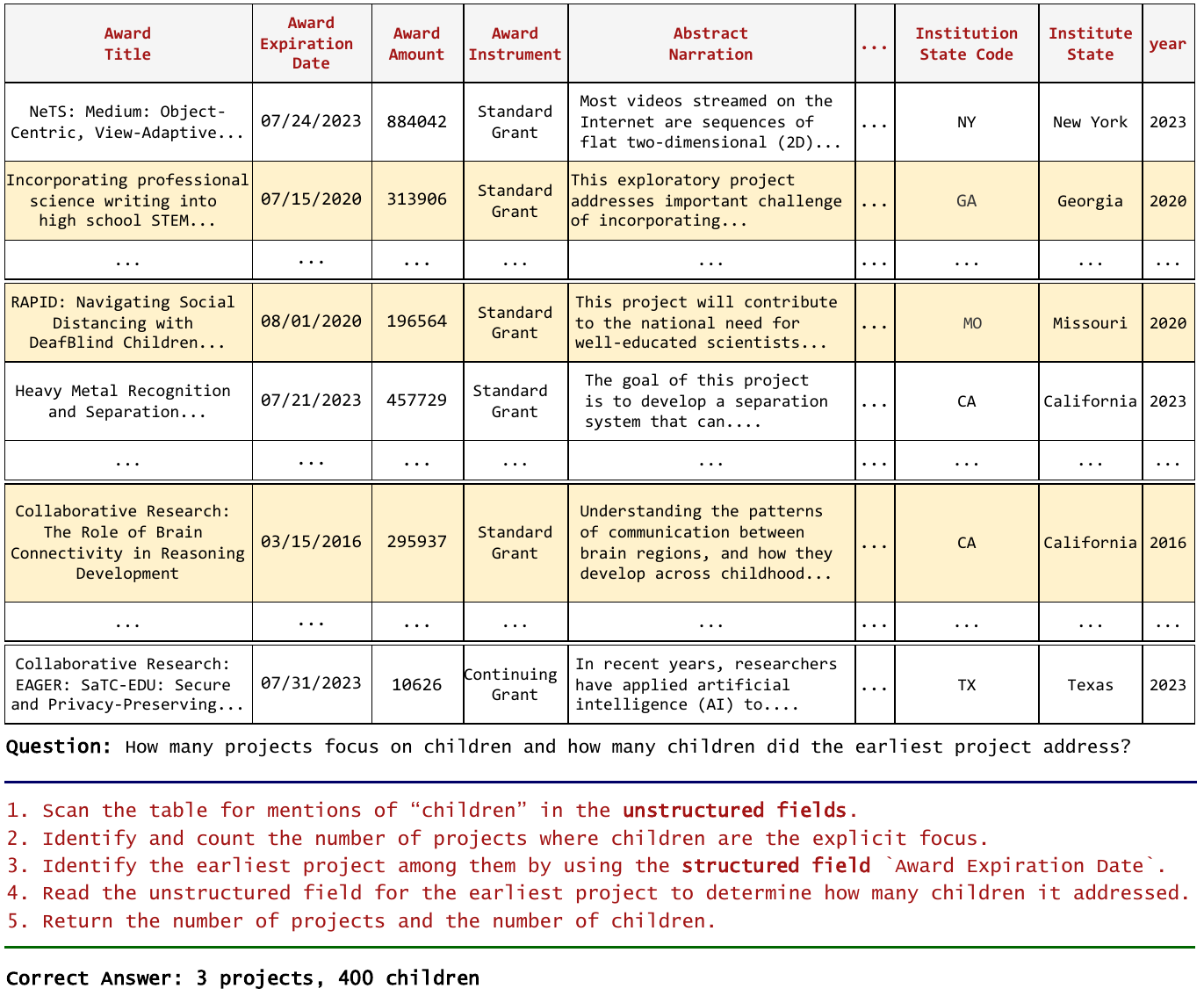}
  \vspace{-1em}
  \caption{\textbf{Example from the RB-Science subset}. The question requires understanding data from \textit{unstructured fields}, aggregation across rows, temporal reasoning to identify the earliest project, and recognition of underspecified information, highlighting challenges beyond surface retrieval.}
  \vspace{-.5em}
  \label{fig:illustraion_2}
\end{figure*}

\begin{figure*}
  \vspace{-0.75em}
  \centering
  \includegraphics[width=\linewidth]{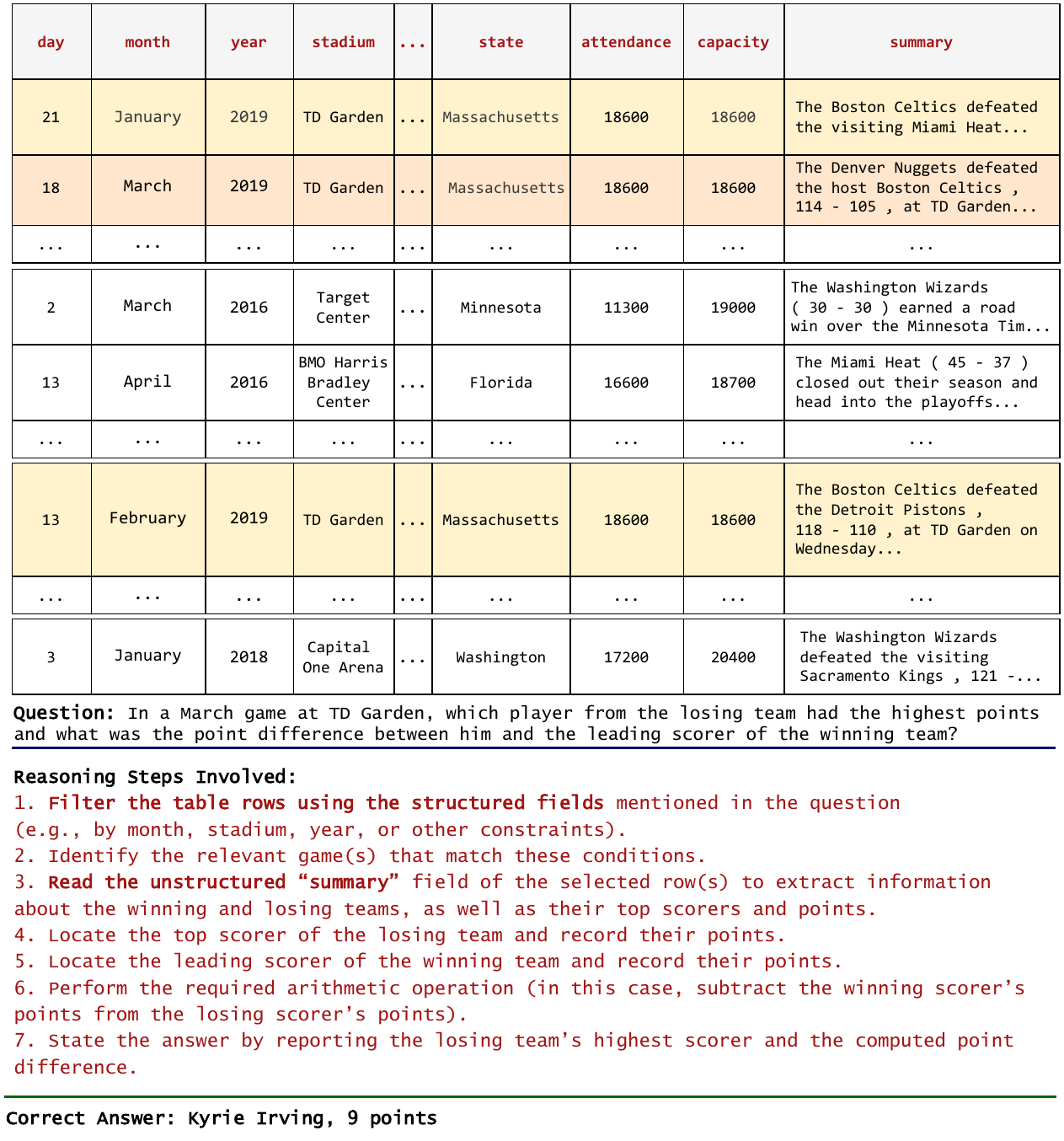}
  \vspace{-1em}
  \caption{\textbf{Example from the RB-Sports subset}. Answering the question requires filtering by \textit{structured fields} (month, stadium), extracting top scorers from \textit{unstructured summaries}, and performing arithmetic comparison, illustrating hybrid multi-hop reasoning across modalities.}
  \vspace{-.5em}
  \label{fig:illustraion_8}
\end{figure*}

Semi-structured tables in \AlgName pose a unique challenge for LLMs, as they require reasoning that spans both structured schema elements (e.g., categorical or numeric fields) and unstructured text (e.g., summaries or descriptions). Such inputs expose the limitations of models that excel in either symbolic precision or semantic understanding, but not both. As illustrated in Figure~\ref{fig:illustraion_2}, answering 
\texttt{`How many projects focus on children and how many children did the earliest project address?'} requires scanning abstracts for child-related projects, counting across rows, and applying temporal reasoning to identify the earliest award. Crucially, the abstract of the 2016 brain connectivity project mentions developmental trajectories without specifying participant numbers, so the correct response must acknowledge the absence of detail. Similarly, for the question \texttt{`In a March game at TD Garden, which player from the losing team had the highest points and what was the point difference between him and the leading scorer of the winning team?'} (Figure~\ref{fig:illustraion_8}) requires filtering structured fields to locate the relevant March 2019 Celtics–Nuggets game, extracting top scorers from the unstructured summary, aligning them with their teams, and performing arithmetic to compute the score difference. This case exemplifies hybrid reasoning across structured and unstructured inputs, combined with entity disambiguation and grounded numeric comparison. These cases underscore how \AlgName questions move beyond single-field lookup, requiring schema filtering, semantic interpretation, aggregation, and handling capabilities that remain fragile in current LLMs.

\section{Error Analysis}
\label{sec:error}
To analyze the sources of performance degradation, we manually examined 100 randomly sampled erroneous predictions from \texttt{Gemini-2.0-Flash (CoT)}. Errors were grouped into four major categories reflecting distinct failure modes: (i) \textbf{Interpretation Error:} counting or lookup mistakes caused by complex table structures and increased token load from unstructured fields; (ii) \textbf{Logical Inconsistency Errors:} contradictory or incomplete reasoning chains, particularly in multi-hop settings; (iii) \textbf{Misalignment Errors:} outputs that deviate from the expected answer schema or provide only partial results; and (iv) \textbf{Extraction Errors}: incorrect or missed retrievals from structured or unstructured regions of the table. The breakdown in Table~\ref{tab:error-breakdown} shows that no single type dominates; instead, errors stem from the interaction between structural complexity, multi-step reasoning, and representational inconsistencies introduced by semi-structured inputs.

\begin{figure*}
  \vspace{-0.75em}
  \centering
  \includegraphics[width=\linewidth]{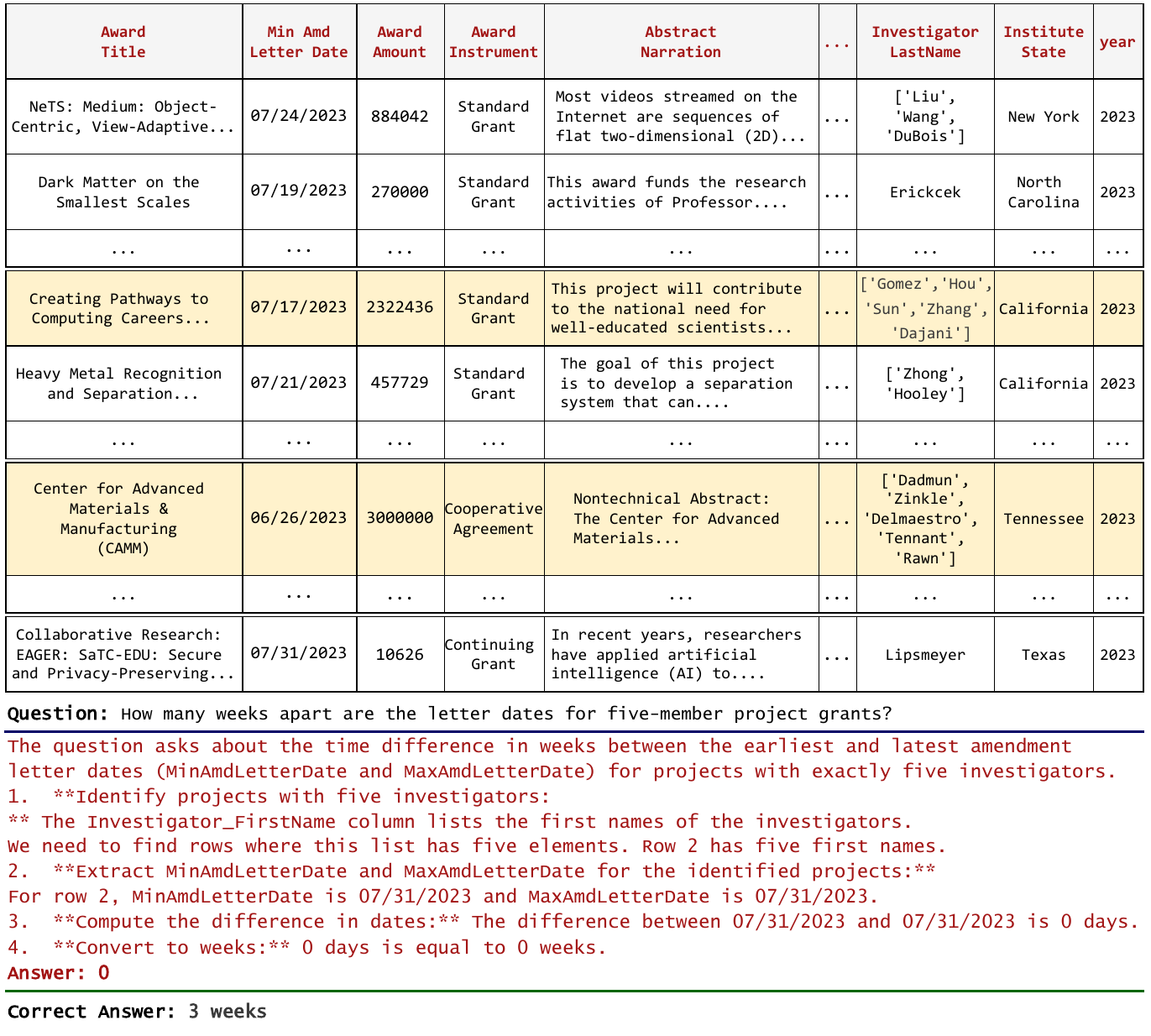}
  \vspace{-1em}
  \caption{\textbf{Extraction Error.} The LLM fails to extract the relevant information from the structured table. Instead of identifying the number of projects sanctioned in October 2022 and comparing it with October 2023, it wrongly concludes that no “previous October” exists.}
  \vspace{-.5em}
  \label{fig:illustration_3}
\end{figure*}

\begin{figure*}
  \vspace{-0.75em}
  \centering
  \includegraphics[width=\linewidth]{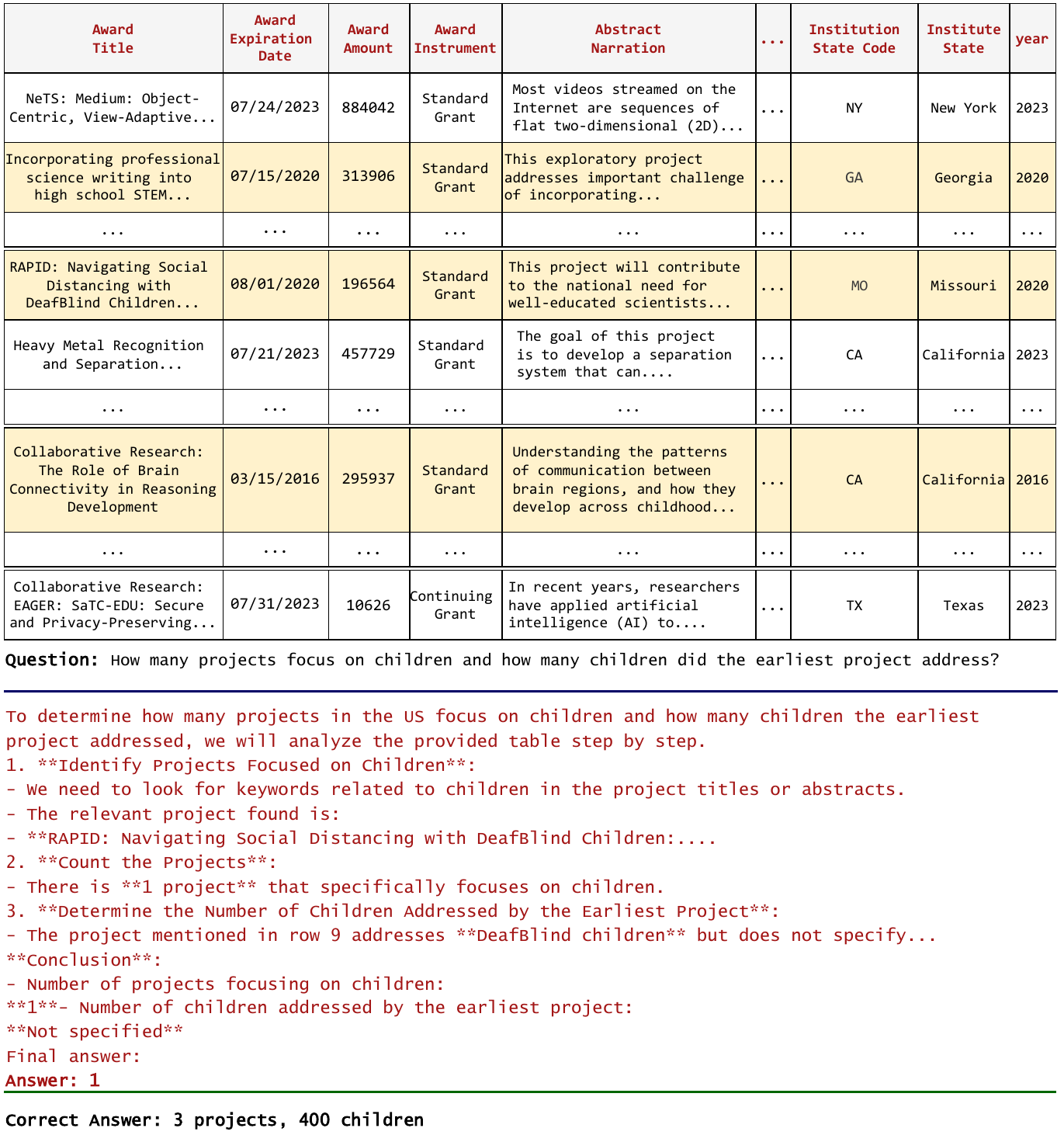}
  \vspace{-1em}
  \caption{\textbf{Extraction Error.} The LLM fails to extract relevant information from the unstructured portion of the table. While only one project explicitly mentions the term \textit{children} in its title, two additional projects are related but require a deeper comprehension of the unstructured content to be correctly identified and extracted.}
  \vspace{-.5em}
  \label{fig:illustration_7}
\end{figure*}

\begin{figure*}
  \vspace{-0.75em}
  \centering
  \includegraphics[width=\linewidth]{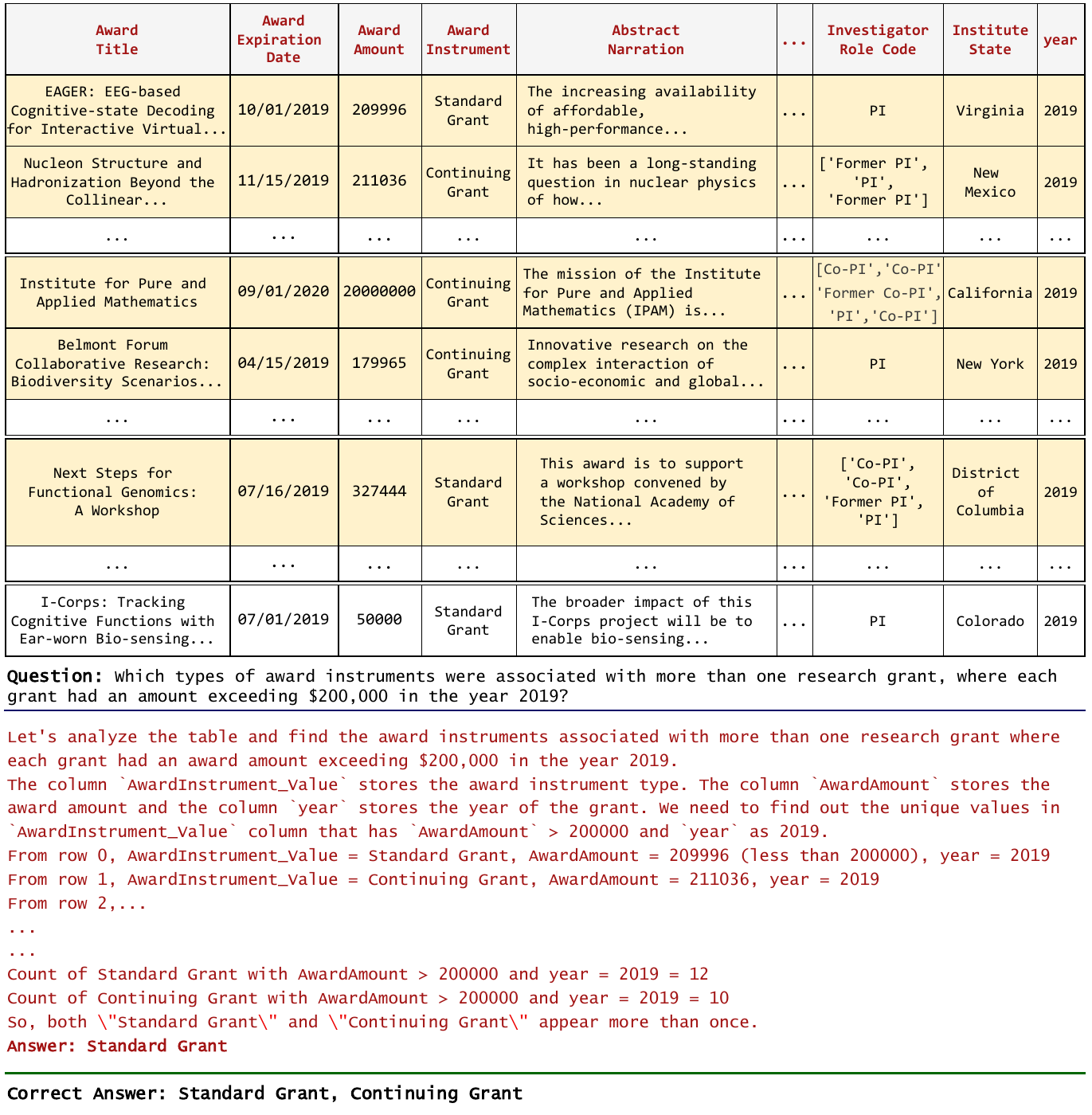}
  \vspace{-1em}
  \caption{\textbf{Logical Inconsistency Error.} Owing to the large number of rows, the LLM engages in extensive reasoning and correctly identifies both \textit{Standard Grant} and \textit{Continuing Grant}. However, the final answer only lists Standard Grant, revealing a collapse between reasoning and output under heavy analysis. }
  \vspace{-.5em}
  \label{fig:illustration_4}
\end{figure*}

\begin{figure*}
  \vspace{-0.75em}
  \centering
  \includegraphics[width=\linewidth]{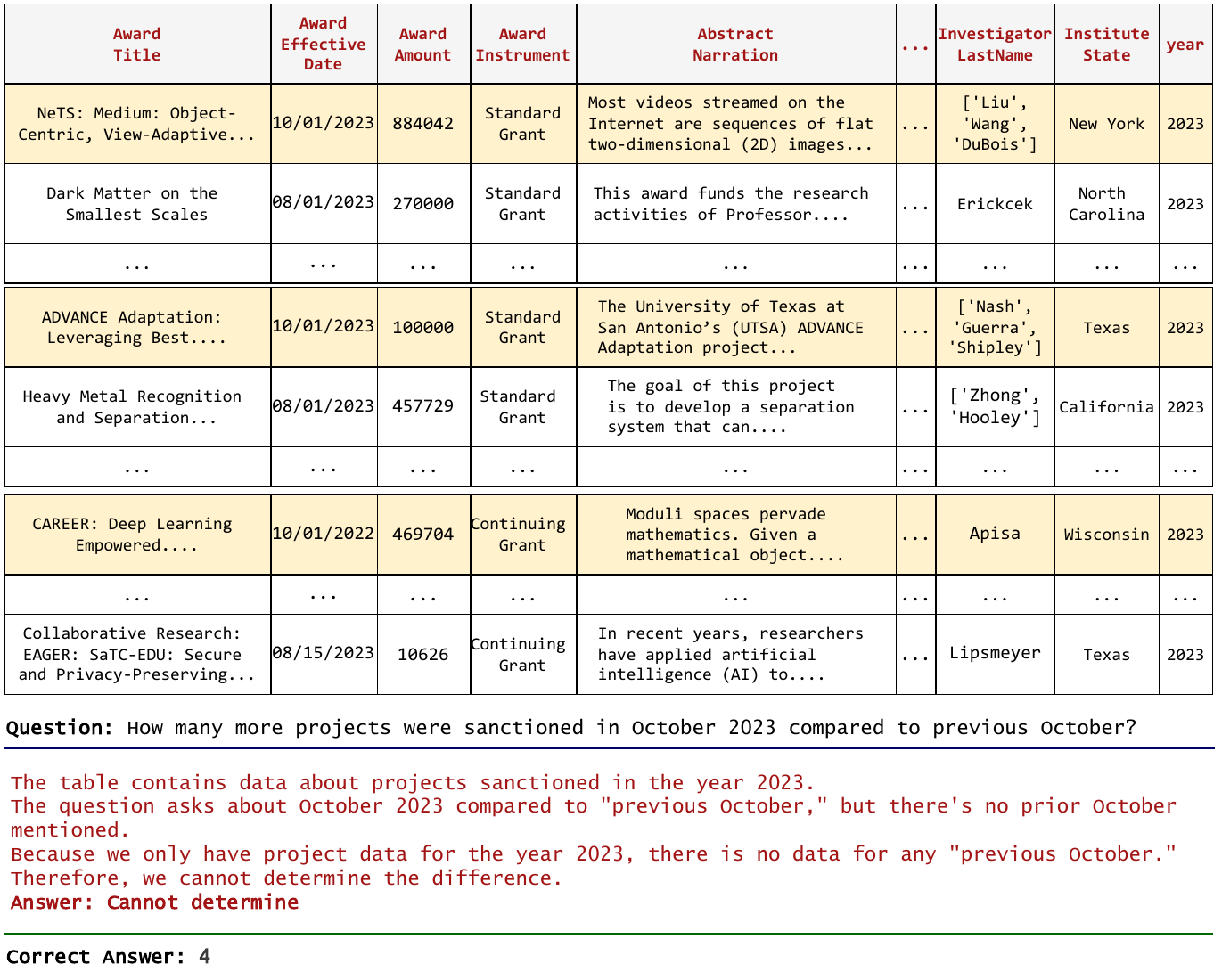}
  \vspace{-1em}
  \caption{\textbf{Interpretation Error}, where the LLM misinterprets both the question and the table. While tasked with finding the week gap between the earliest and latest amendment dates for five-investigator projects, it only considers a single row and ignores other valid rows. This leads to an incorrect calculation, showing how errors in interpreting table structure and question scope can cascade into a wrong final answer.}
  \vspace{-.5em}
  \label{fig:illustration_5}
\end{figure*}

\begin{figure*}
  \vspace{-0.75em}
  \centering
  \includegraphics[width=\linewidth]{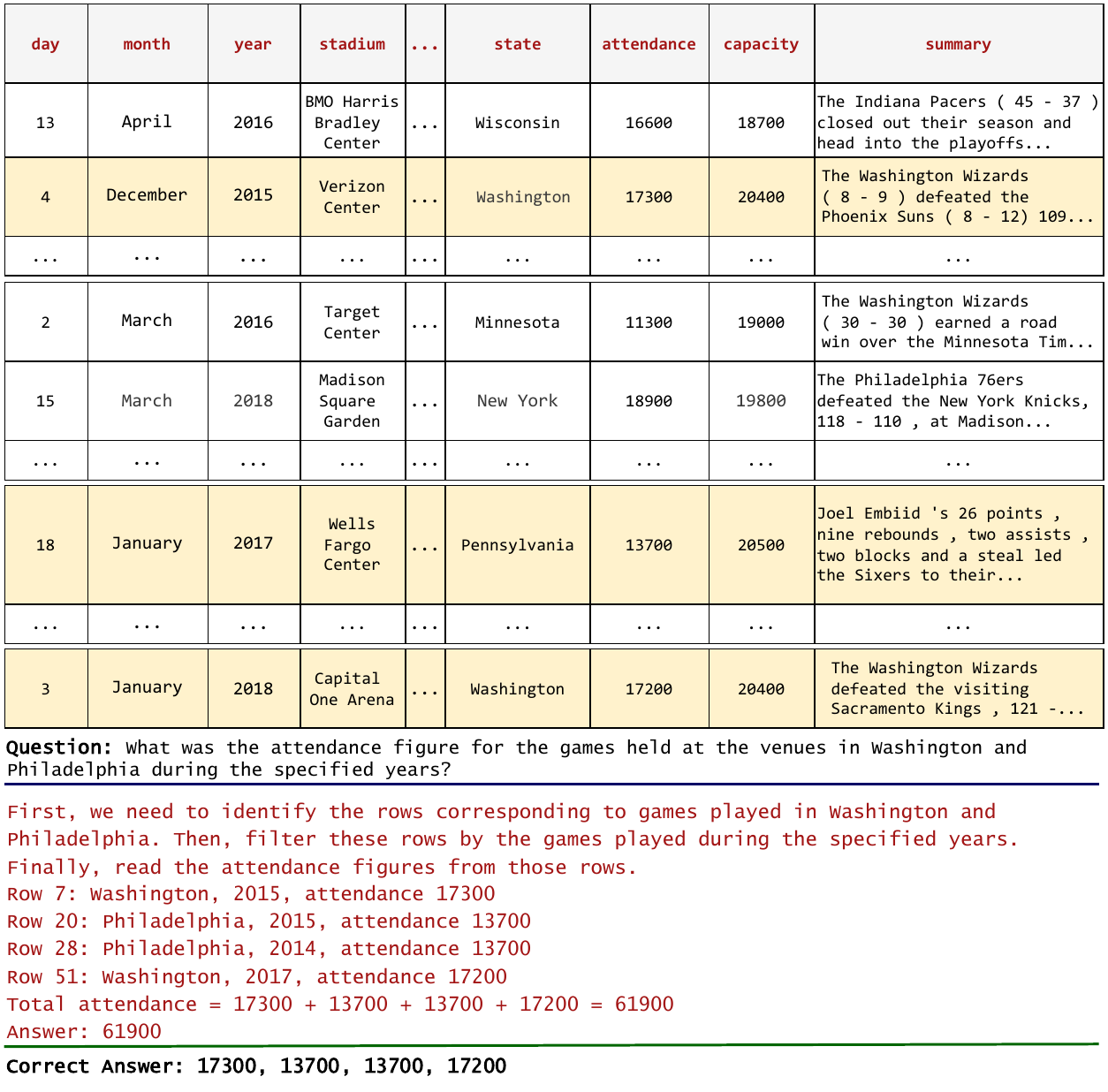}
  \vspace{-1em}
  \caption{\textbf{Misalignment Error.} The reasoning correctly identifies the relevant rows and extracts the attendance figures. However, instead of listing these individual values as expected, the LLM sums them up. This misalignment between the required output format and the final answer leads to an incorrect response despite accurate intermediate reasoning.}
  \vspace{-.5em}
  \label{fig:illustration_6}
\end{figure*}

\begin{table}[!htbp]
\centering
\vspace{-.5em}
\caption{\small Breakdown of 100 randomly sampled erroneous predictions from \texttt{Gemini-2.0-Flash (CoT)}.}
\setlength{\tabcolsep}{5pt}
\fontsize{10}{11}\selectfont
\label{tab:error-breakdown}
\begin{tabular}{lr}
    \toprule
    \textbf{Error Type} & \textbf{Percentage} \\
    \midrule
    Interpretation Error & 22\% \\
    Logical Inconsistencies & 31\% \\
    Misalignment Error & 27\% \\
    Extraction Error & 20\% \\
\bottomrule
\bottomrule
\end{tabular}
\end{table}
\vspace{-0.75em}
\paragraph{Extraction Error.}
These involve failures to retrieve key information from structured fields or unstructured text. The model may skip valid rows or miss implicit cues, such as differences in project counts across years (Figure~\ref{fig:illustration_3}) or mentions of child-related studies buried in abstracts (Figure~\ref{fig:illustration_7}).
% \vspace{-0.75em}
\paragraph{Logical Inconsistency Error.}
These occur when the model generates an apparently coherent reasoning chain but produces a final answer inconsistent with its intermediate analysis. For example, as shown in Figure~\ref{fig:illustration_4}, the model may identify both \textit{Standard Grant} and \textit{Continuing Grant} as valid answers but report only one, revealing a collapse between reasoning and final output generation.
% \vspace{-0.75em}
\paragraph{Interpretation Error.}
Here, the model misreads the scope of the question or the table structure, overlooking relevant rows or applying filters incorrectly. As illustrated in Figure~\ref{fig:illustration_5}, it may compute time gaps based on a single record while ignoring other valid entries, leading to incomplete evidence gathering and erroneous conclusions.
% \vspace{-0.75em}
% \vspace{-1.25em}
\paragraph{Misalignment Error.}
In some cases, the model’s reasoning is correct, but the output format deviates from the expected answer schema—for instance, returning a sum instead of individual attendance values (Figure~\ref{fig:illustration_6}). Collectively, these patterns show that while LLMs can perform multi-step reasoning, they often lose alignment between reasoning, evidence retrieval, and output generation particularly when operating on semi-structured data that demands both symbolic precision and semantic understanding.

%% file: custom.bib
@article{nan2022fetaqa,
  title={FeTaQA: Free-form table question answering},
  author={Nan, Linyong and Hsieh, Chiachun and Mao, Ziming and Lin, Xi Victoria and Verma, Neha and Zhang, Rui and Kry{\'s}ci{\'n}ski, Wojciech and Schoelkopf, Hailey and Kong, Riley and Tang, Xiangru and others},
  journal={Transactions of the Association for Computational Linguistics},
  volume={10},
  pages={35--49},
  year={2022},
  publisher={MIT Press One Broadway, 12th Floor, Cambridge, Massachusetts 02142, USA~…}
}

@article{abhyankar2024h,
  title={H-STAR: LLM-driven Hybrid SQL-Text Adaptive Reasoning on Tables},
  author={Abhyankar, Nikhil and Gupta, Vivek and Roth, Dan and Reddy, Chandan K},
  journal={arXiv preprint arXiv:2407.05952},
  year={2024}
}

@inproceedings{gupta2020infotabs,
  title={INFOTABS: Inference on Tables as Semi-structured Data},
  author={Gupta, Vivek and Mehta, Maitrey and Nokhiz, Pegah and Srikumar, Vivek},
  booktitle={Proceedings of the 58th Annual Meeting of the Association for Computational Linguistics},
  pages={2309--2324},
  year={2020}
}

@inproceedings{gupta2023temptabqa,
  title={TempTabQA: Temporal Question Answering for Semi-Structured Tables},
  author={Gupta, Vivek and Kandoi, Pranshu and Vora, Mahek and Zhang, Shuo and He, Yujie and Reinanda, Ridho and Srikumar, Vivek},
  booktitle={Proceedings of the 2023 Conference on Empirical Methods in Natural Language Processing},
  pages={2431--2453},
  year={2023}
}

@inproceedings{iyyer2017search,
  title={Search-based neural structured learning for sequential question answering},
  author={Iyyer, Mohit and Yih, Wen-tau and Chang, Ming-Wei},
  booktitle={Proceedings of the 55th Annual Meeting of the Association for Computational Linguistics (Volume 1: Long Papers)},
  pages={1821--1831},
  year={2017}
}

@inproceedings{yu2018spider,
  title={Spider: A Large-Scale Human-Labeled Dataset for Complex and Cross-Domain Semantic Parsing and Text-to-SQL Task},
  author={Yu, Tao and Zhang, Rui and Yang, Kai and Yasunaga, Michihiro and Wang, Dongxu and Li, Zifan and Ma, James and Li, Irene and Yao, Qingning and Roman, Shanelle and others},
  booktitle={Proceedings of the 2018 Conference on Empirical Methods in Natural Language Processing},
  pages={3911--3921},
  year={2018}
}

@inproceedings{aly2021fact,
  title={The fact extraction and VERification over unstructured and structured information (FEVEROUS) shared task},
  author={Aly, Rami and Guo, Zhijiang and Schlichtkrull, Michael Sejr and Thorne, James and Vlachos, Andreas and Christodoulopoulos, Christos and Cocarascu, Oana and Mittal, Arpit},
  booktitle={Proceedings of the Fourth Workshop on Fact Extraction and VERification (FEVER)},
  pages={1--13},
  year={2021}
}

@article{zhong2017seq2sql,
  title={Seq2sql: Generating structured queries from natural language using reinforcement learning},
  author={Zhong, Victor and Xiong, Caiming and Socher, Richard},
  journal={arXiv preprint arXiv:1709.00103},
  year={2017}
}

@inproceedings{chen2020hybridqa,
  title={HybridQA: A Dataset of Multi-Hop Question Answering over Tabular and Textual Data},
  author={Chen, Wenhu and Zha, Hanwen and Chen, Zhiyu and Xiong, Wenhan and Wang, Hong and Wang, William Yang},
  booktitle={Findings of the Association for Computational Linguistics: EMNLP 2020},
  pages={1026--1036},
  year={2020}
}

@inproceedings{chen2021finqa,
  title={FinQA: A Dataset of Numerical Reasoning over Financial Data},
  author={Chen, Zhiyu and Chen, Wenhu and Smiley, Charese and Shah, Sameena and Borova, Iana and Langdon, Dylan and Moussa, Reema and Beane, Matt and Huang, Ting-Hao and Routledge, Bryan R and others},
  booktitle={Proceedings of the 2021 Conference on Empirical Methods in Natural Language Processing},
  pages={3697--3711},
  year={2021}
}

@inproceedings{zhu2021tat,
  title={TAT-QA: A Question Answering Benchmark on a Hybrid of Tabular and Textual Content in Finance},
  author={Zhu, Fengbin and Lei, Wenqiang and Huang, Youcheng and Wang, Chao and Zhang, Shuo and Lv, Jiancheng and Feng, Fuli and Chua, Tat-Seng},
  booktitle={Proceedings of the 59th Annual Meeting of the Association for Computational Linguistics and the 11th International Joint Conference on Natural Language Processing (Volume 1: Long Papers)},
  pages={3277--3287},
  year={2021}
}

@inproceedings{zhang2023crt,
  title={CRT-QA: A dataset of complex reasoning question answering over tabular data},
  author={Zhang, Zhehao and Li, Xitao and Gao, Yan and Lou, Jian-Guang},
  booktitle={Proceedings of the 2023 Conference on Empirical Methods in Natural Language Processing},
  pages={2131--2153},
  year={2023}
}

@inproceedings{lu2023scitab,
  title={SCITAB: A Challenging Benchmark for Compositional Reasoning and Claim Verification on Scientific Tables},
  author={Lu, Xinyuan and Pan, Liangming and Liu, Qian and Nakov, Preslav and Kan, Min-Yen},
  booktitle={Proceedings of the 2023 Conference on Empirical Methods in Natural Language Processing},
  pages={7787--7813},
  year={2023}
}

@inproceedings{wang2021semeval,
  title={SemEval-2021 Task 9: Fact Verification and Evidence Finding for Tabular Data in Scientific Documents (SEM-TAB-FACTS)},
  author={Wang, Nancy XR and Mahajan, Diwakar and Danilevsky, Marina and Rosenthal, Sara},
  booktitle={Proceedings of the 15th International Workshop on Semantic Evaluation (SemEval-2021)},
  pages={317--326},
  year={2021}
}

@inproceedings{wang2023chain,
  title={Chain-of-Table: Evolving Tables in the Reasoning Chain for Table Understanding},
  author={Wang, Zilong and Zhang, Hao and Li, Chun-Liang and Eisenschlos, Julian Martin and Perot, Vincent and Wang, Zifeng and Miculicich, Lesly and Fujii, Yasuhisa and Shang, Jingbo and Lee, Chen-Yu and others},
  booktitle={The Twelfth International Conference on Learning Representations},
  year={2023}
}

@article{nahid2024tabsqlify,
  title={TabSQLify: Enhancing Reasoning Capabilities of LLMs Through Table Decomposition},
  author={Nahid, Md Mahadi Hasan and Rafiei, Davood},
  journal={arXiv preprint arXiv:2404.10150},
  year={2024}
}

@article{chen2023program,
  title={Program of Thoughts Prompting: Disentangling Computation from Reasoning for Numerical Reasoning Tasks},
  author={Chen, Wenhu and Ma, Xueguang and Wang, Xinyi and Cohen, William W},
  journal={Transactions on Machine Learning Research},
  year={2023}
}

@inproceedings{cheng2022binding,
  title={Binding Language Models in Symbolic Languages},
  author={Cheng, Zhoujun and Xie, Tianbao and Shi, Peng and Li, Chengzu and Nadkarni, Rahul and Hu, Yushi and Xiong, Caiming and Radev, Dragomir and Ostendorf, Mari and Zettlemoyer, Luke and others},
  booktitle={The Eleventh International Conference on Learning Representations},
  year={2022}
}

@inproceedings{chen2019tabfact,
  title={TabFact: A Large-scale Dataset for Table-based Fact Verification},
  author={Chen, Wenhu and Wang, Hongmin and Chen, Jianshu and Zhang, Yunkai and Wang, Hong and Li, Shiyang and Zhou, Xiyou and Wang, William Yang},
  booktitle={International Conference on Learning Representations},
  year={2019}
}

@inproceedings{pasupat2015compositional,
  title={Compositional Semantic Parsing on Semi-Structured Tables},
  author={Pasupat, Panupong and Liang, Percy},
  booktitle={Proceedings of the 53rd Annual Meeting of the Association for Computational Linguistics and the 7th International Joint Conference on Natural Language Processing (Volume 1: Long Papers)},
  pages={1470--1480},
  year={2015}
}

@article{openai2023gpt,
  title={Gpt-4 technical report. arxiv 2303.08774},
  author={OpenAI},
  journal={View in Article},
  volume={2},
  number={5},
  year={2023}
}

@article{team2023gemini,
  title={Gemini: a family of highly capable multimodal models},
  author={Team, Gemini and Anil, Rohan and Borgeaud, Sebastian and Alayrac, Jean-Baptiste and Yu, Jiahui and Soricut, Radu and Schalkwyk, Johan and Dai, Andrew M and Hauth, Anja and Millican, Katie and others},
  journal={arXiv preprint arXiv:2312.11805},
  year={2023}
}

@inproceedings{chen2023large,
  title={Large Language Models are few (1)-shot Table Reasoners},
  author={Chen, Wenhu},
  booktitle={Findings of the Association for Computational Linguistics: EACL 2023},
  pages={1120--1130},
  year={2023}
}

@inproceedings{chen2020open,
  title={Open Question Answering over Tables and Text},
  author={Chen, Wenhu and Chang, Ming-Wei and Schlinger, Eva and Wang, William Yang and Cohen, William W},
  booktitle={International Conference on Learning Representations},
  year={2020}
}

@article{liu2023rethinking,
  title={Rethinking Tabular Data Understanding with Large Language Models},
  author={Liu, Tianyang and Wang, Fei and Chen, Muhao},
  journal={arXiv preprint arXiv:2312.16702},
  year={2023}
}

@article{dubey2024llama,
  title={The llama 3 herd of models},
  author={Dubey, Abhimanyu and Jauhri, Abhinav and Pandey, Abhinav and Kadian, Abhishek and Al-Dahle, Ahmad and Letman, Aiesha and Mathur, Akhil and Schelten, Alan and Yang, Amy and Fan, Angela and others},
  journal={arXiv preprint arXiv:2407.21783},
  year={2024}
}

@article{lu2022dynamic,
  title={Dynamic prompt learning via policy gradient for semi-structured mathematical reasoning},
  author={Lu, Pan and Qiu, Liang and Chang, Kai-Wei and Wu, Ying Nian and Zhu, Song-Chun and Rajpurohit, Tanmay and Clark, Peter and Kalyan, Ashwin},
  journal={arXiv preprint arXiv:2209.14610},
  year={2022}
}

@inproceedings{zhao2022multihiertt,
  title={MultiHiertt: Numerical Reasoning over Multi Hierarchical Tabular and Textual Data},
  author={Zhao, Yilun and Li, Yunxiang and Li, Chenying and Zhang, Rui},
  booktitle={Proceedings of the 60th Annual Meeting of the Association for Computational Linguistics (Volume 1: Long Papers)},
  pages={6588--6600},
  year={2022}
}

@inproceedings{thomson2020sportsett,
  title={SportSett: basketball-a robust and maintainable data-set for natural language generation},
  author={Thomson, Craig and Reiter, Ehud and Sripada, Somayajulu},
  booktitle={Proceedings of the Workshop on Intelligent Information Processing and Natural Language Generation},
  pages={32--40},
  year={2020}
}

@inproceedings{moosavi2021scigen,
  title={Scigen: a dataset for reasoning-aware text generation from scientific tables},
  author={Moosavi, Nafise Sadat and R{\"u}ckl{\'e}, Andreas and Roth, Dan and Gurevych, Iryna},
  booktitle={Thirty-fifth Conference on Neural Information Processing Systems Datasets and Benchmarks Track (Round 2)},
  year={2021}
}

@inproceedings{zheng2021global,
  title={Global table extractor (gte): A framework for joint table identification and cell structure recognition using visual context},
  author={Zheng, Xinyi and Burdick, Douglas and Popa, Lucian and Zhong, Xu and Wang, Nancy Xin Ru},
  booktitle={Proceedings of the IEEE/CVF winter conference on applications of computer vision},
  pages={697--706},
  year={2021}
}

@article{wu2024protrix,
  title={ProTrix: building models for planning and reasoning over tables with sentence context},
  author={Wu, Zirui and Feng, Yansong},
  journal={arXiv preprint arXiv:2403.02177},
  year={2024}
}

@article{glenn2024blendsql,
  title={BlendSQL: A scalable dialect for unifying hybrid question answering in relational algebra},
  author={Glenn, Parker and Dakle, Parag Pravin and Wang, Liang and Raghavan, Preethi},
  journal={arXiv preprint arXiv:2402.17882},
  year={2024}
}

@inproceedings{park2023generative,
  title={Generative agents: Interactive simulacra of human behavior},
  author={Park, Joon Sung and O'Brien, Joseph and Cai, Carrie Jun and Morris, Meredith Ringel and Liang, Percy and Bernstein, Michael S},
  booktitle={Proceedings of the 36th annual acm symposium on user interface software and technology},
  pages={1--22},
  year={2023}
}

@inproceedings{li2024planning,
  title={Planning first, question second: An LLM-guided method for controllable question generation},
  author={Li, Kunze and Zhang, Yu},
  booktitle={Findings of the Association for Computational Linguistics ACL 2024},
  pages={4715--4729},
  year={2024}
}

@article{wiseman2017challenges,
  title={Challenges in data-to-document generation},
  author={Wiseman, Sam and Shieber, Stuart M and Rush, Alexander M},
  journal={arXiv preprint arXiv:1707.08052},
  year={2017}
}

@misc{NSF2024,
  author       = {NSF},
  title        = {National Science Foundation(NSF)},
  year         = {2024},
  url          = {https://www.nsf.gov/example-page},
}

@misc{Wikipedia,
  author       = {Wikipedia},
  title        = {Wikipedia},
  url          = {https://www.wikipedia.org/},
}

@misc{Financial,
  author       = {Financial Reports},
  title        = {Annual Reports},
  year         = {2025},
  url          = {https://www.annualreports.com/},
}

@article{srivastava2022beyond,
  title={Beyond the imitation game: Quantifying and extrapolating the capabilities of language models},
  author={Srivastava, Aarohi and Rastogi, Abhinav and Rao, Abhishek and Shoeb, Abu Awal Md and Abid, Abubakar and Fisch, Adam and Brown, Adam R and Santoro, Adam and Gupta, Aditya and Garriga-Alonso, Adri{\`a} and others},
  journal={arXiv preprint arXiv:2206.04615},
  year={2022}
}

@article{yang2025qwen3,
  title={Qwen3 technical report},
  author={Yang, An and Li, Anfeng and Yang, Baosong and Zhang, Beichen and Hui, Binyuan and Zheng, Bo and Yu, Bowen and Gao, Chang and Huang, Chengen and Lv, Chenxu and others},
  journal={arXiv preprint arXiv:2505.09388},
  year={2025}
}

@article{guo2025deepseek,
  title={Deepseek-r1: Incentivizing reasoning capability in llms via reinforcement learning},
  author={Guo, Daya and Yang, Dejian and Zhang, Haowei and Song, Junxiao and Zhang, Ruoyu and Xu, Runxin and Zhu, Qihao and Ma, Shirong and Wang, Peiyi and Bi, Xiao and others},
  journal={arXiv preprint arXiv:2501.12948},
  year={2025}
}

@article{wei2022chain,
  title={Chain-of-thought prompting elicits reasoning in large language models},
  author={Wei, Jason and Wang, Xuezhi and Schuurmans, Dale and Bosma, Maarten and Xia, Fei and Chi, Ed and Le, Quoc V and Zhou, Denny and others},
  journal={Advances in neural information processing systems},
  volume={35},
  pages={24824--24837},
  year={2022}
}

@article{liu2023lost,
  title={Lost in the middle: How language models use long contexts},
  author={Liu, Nelson F and Lin, Kevin and Hewitt, John and Paranjape, Ashwin and Bevilacqua, Michele and Petroni, Fabio and Liang, Percy},
  journal={arXiv preprint arXiv:2307.03172},
  year={2023}
}

@inproceedings{papineni2002bleu,
  title={Bleu: a method for automatic evaluation of machine translation},
  author={Papineni, Kishore and Roukos, Salim and Ward, Todd and Zhu, Wei-Jing},
  booktitle={Proceedings of the 40th annual meeting of the Association for Computational Linguistics},
  pages={311--318},
  year={2002}
}

@misc{jiang2024mixtral,
  title        = {Mistral-Small-3.2-24B-Instruct-2506},
  author       = {Mistral},
  howpublished = {\url{https://huggingface.co/mistralai/Mistral-Small-3.2-24B-Instruct-2506}},
  note         = {Hugging Face model card; 24B parameters; updated version of Small-3.1},
  year         = {2024}
}

@article{nahid2024normtab,
  title={NormTab: Improving symbolic reasoning in LLMs through tabular data normalization},
  author={Nahid, Md Mahadi Hasan and Rafiei, Davood},
  journal={arXiv preprint arXiv:2406.17961},
  year={2024}
}

@article{cao2025tablemaster,
  title={Tablemaster: A recipe to advance table understanding with language models},
  author={Cao, Lang and Liu, Hanbing},
  journal={arXiv preprint arXiv:2501.19378},
  year={2025}
}
